\documentclass{article}

\usepackage{microtype}
\usepackage{graphicx}
\usepackage{subcaption}
\usepackage{booktabs} % for professional tables
\usepackage{multirow}

\usepackage{hyperref}

\usepackage[preprint]{icml2026}

\usepackage{amsmath}
\usepackage{amssymb}
\usepackage{mathtools}
\usepackage{amsthm}

\usepackage[capitalize,noabbrev]{cleveref}

\theoremstyle{plain}

\theoremstyle{definition}

\theoremstyle{remark}

\usepackage[textsize=tiny]{todonotes}

\icmltitlerunning{Semi-Supervised Prototypical Networks for Few-Shot Class Incremental Learning}

\begin{document}

\twocolumn[
    \icmltitle{SPRINT: Semi-supervised Prototypical Representation for Few-Shot Class-Incremental Tabular Learning}



  \icmlsetsymbol{equal}{*}

    \begin{icmlauthorlist}
      \icmlauthor{Umid Suleymanov}{vt}
      \icmlauthor{Murat Kantarcioglu}{vt,utd}
      \icmlauthor{Kevin S Chan}{arl}
      \icmlauthor{Michael De Lucia}{arl}
      \icmlauthor{Kevin Hamlen}{utd}
      \icmlauthor{Latifur Khan}{utd}
      \icmlauthor{Sharad Mehrotra}{uci}
      \icmlauthor{Ananthram Swami}{arl}
      \icmlauthor{Bhavani Thuraisingham}{utd}
    \end{icmlauthorlist}
    
    \icmlaffiliation{vt}{Department of Computer Science, Virginia Tech, Blacksburg, VA, USA}
    \icmlaffiliation{arl}{DEVCOM Army Research Laboratory, Adelphi, MD, USA}
    \icmlaffiliation{utd}{Department of Computer Science, University of Texas at Dallas, Dallas, TX, USA}
    \icmlaffiliation{uci}{Department of Computer Science, University of California, Irvine, CA, USA}
    
    \icmlcorrespondingauthor{Umid Suleymanov}{umids@vt.edu}

  \icmlkeywords{Machine Learning, ICML}

  \vskip 0.3in
]



\printAffiliationsAndNotice{}  % no special notice (required even if empty)

\begin{abstract}

%MK-Done: add a sentence or two to motivate FSCIL first.
%MK-Done: what we do make it tailored for tabular data?
Real-world systems must continuously adapt to novel concepts from limited data without forgetting previously acquired knowledge. While Few-Shot Class-Incremental Learning (FSCIL) is established in computer vision, its application to \textbf{tabular domains} remains largely unexplored. Unlike images, tabular streams (e.g., logs, sensors) offer abundant unlabeled data, a scarcity of expert annotations and negligible storage costs – features ignored by existing vision-based methods that rely on restrictive buffers. We introduce \textbf{SPRINT}, the first FSCIL framework tailored for tabular distributions. SPRINT introduces a mixed episodic training strategy that leverages confidence-based pseudo-labeling to enrich novel class representations and exploits low storage costs to retain base class history.  Extensive evaluation across six diverse benchmarks – spanning \textbf{cybersecurity, healthcare, and ecological domains} – demonstrates SPRINT's cross-domain robustness. It achieves a state-of-the-art average accuracy of 77.37\% (5-shot), outperforming the strongest incremental baseline by 4.45\%. 

%MK-Done: it may be good to say how well we are doing compared SOTA.
\end{abstract}

%high-stakes applications. 
%Existing vision-centric FSCIL methods operate under assumptions that are ill-suited for tabular data streams: they enforce strict memory buffer constraints (due to high image storage costs) and rely exclusively on supervised learning, ignoring the massive pools of unlabeled data often available in real-world tabular data tasks such as intrusion detection (e.g., most of the network data used for intrusion detection is not labeled).

\section{Introduction}

Few-Shot Class-Incremental Learning (FSCIL) addresses the dual challenge of adapting to novel classes from limited labeled examples while preventing the catastrophic forgetting of previously acquired knowledge. While this paradigm has been extensively studied in computer vision \citep{tao2020few,zhang2021few,9878986}, the literature has largely overlooked tabular domains, despite their ubiquity in many critical applications. Existing vision-centric FSCIL methods rely on assumptions that are poorly suited to tabular data streams. In particular, they impose strict memory buffer constraints driven by the high storage costs of image data, thereby overlooking the large volumes of unlabeled data commonly available in real-world tabular domains such as intrusion detection, where most network traffic data remains unlabeled.

In addition, tabular data possess unique operational characteristics. Unlike high-dimensional images, tabular records (e.g., network logs, sensor readings, medical records) typically consist of compact feature vectors with low dimensionality. This negligible storage footprint renders the strict fixed memory buffer assumption of vision benchmarks unnecessary. Furthermore, tabular data streams in the wild are rarely fully labeled; they are characterized by a scarcity of expert annotations but an abundance of continuously collected unlabeled samples. A prime example is Network Intrusion Detection Systems (NIDS), where security analysts face a constant stream of unlabeled traffic. New attack variants (novel classes) emerge continuously and must be learned from only a handful of identified signatures ($k$-shot), yet the system must maintain robust detection of historical threats without the downtime associated with full batch retraining. Analogous challenges exist in healthcare epidemiology: hospitals continuously aggregate tabular Electronic Health Records. When a novel pathogen emerges, such as a specific COVID-19 variant, confirmed cases are initially scarce ($k$-shot). Yet, the diagnostic model must rapidly adapt to identify the new strain within the massive stream of patient data. Crucially, the adaptation must occur without losing the ability to diagnose established pathologies like pneumonia.

To address these challenges, we introduce SPRINT (\textbf{S}emi-supervised \textbf{P}rototypical \textbf{R}epresentation for \textbf{IN}cremental \textbf{T}abular learning), a semi-supervised prototypical network designed for tabular FSCIL. Our key contributions are:

\begin{itemize}
  %MK-DOne: Please check this claim!
    \item \textbf{First FSCIL framework tailored for tabular data}: 
    %We formalize Few-Shot Class-Incremental Learning for tabular data with full base data retention and unlabeled data pools – a computationally feasible and operationally realistic setting for tabular domain.    
    We formalize Few-Shot Class-Incremental Learning (FSCIL) for tabular data under a setting that allows retention of base data as memory and access to unlabeled data pools, a computationally feasible and operationally realistic scenario for tabular data applications. 
    \item \textbf{Semi-Supervised Prototype Expansion}: We propose an adaptive strategy that utilizes high-confidence unlabeled samples, enriching novel class representations beyond k-shot examples.
    \item \textbf{Mixed Episodic Training}: We introduce a mixed episodic training strategy that simultaneously optimizes base class rehearsal (sampled from the retained history) and semi-supervised novel adaptation. This implicit joint optimization prevents catastrophic forgetting without requiring complex regularization penalties like distillation.
    \item \textbf{State-of-the-Art Stability}: Extensive evaluation across six diverse benchmarks (including cybersecurity, healthcare, and ecological domains) demonstrates SPRINT's superior robustness. On the ACI-IoT-2023, SPRINT achieves 93.63\% final accuracy with a negligible forgetting rate of 2.54\%, significantly outperforming standard Prototypical Networks (12.03\% forgetting) and incremental baselines.
\end{itemize}
\begin{figure*}
    \centering
    \includegraphics[width=0.9\linewidth]{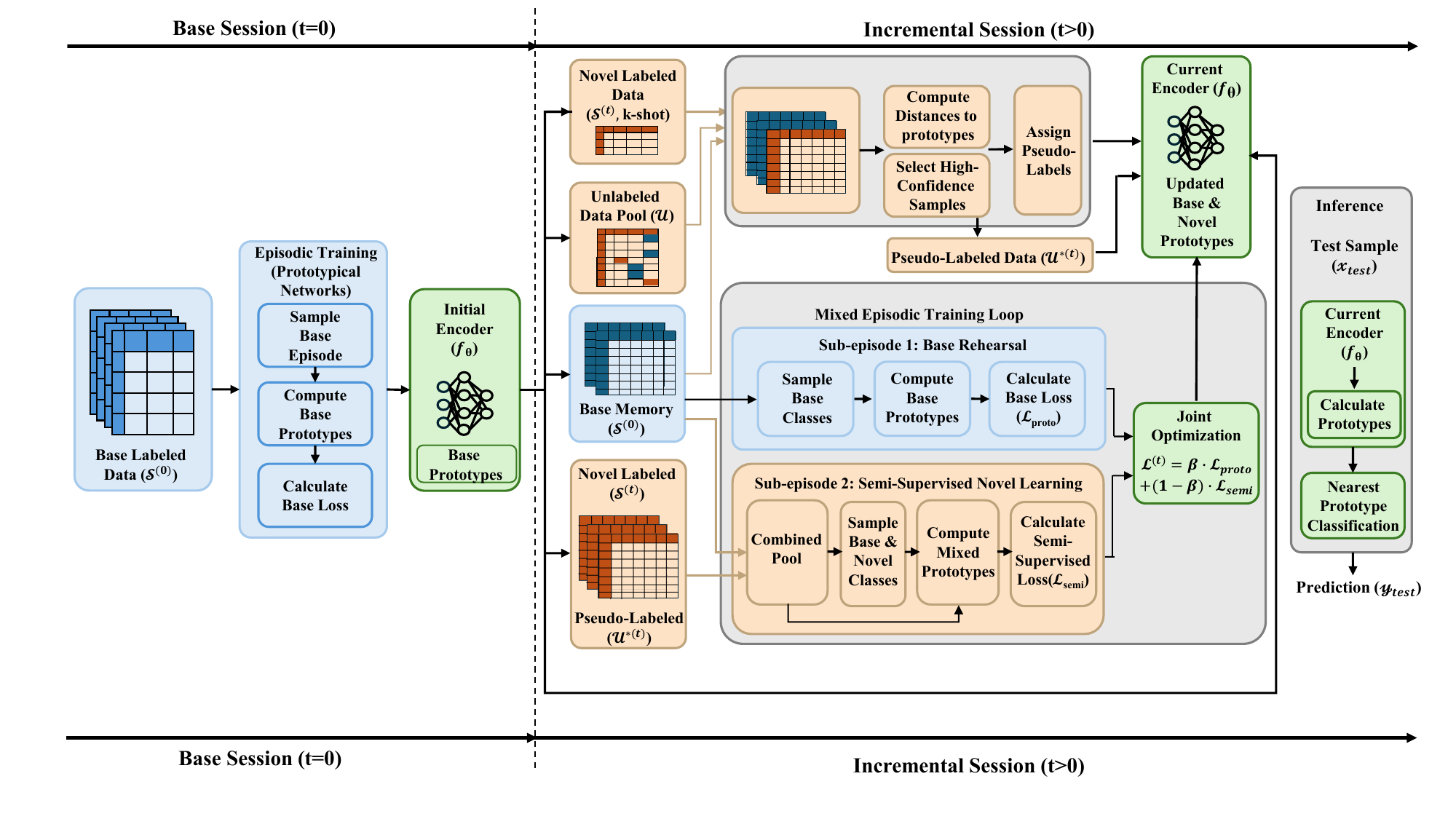}
    %\caption{Method Figure for SPRINT }
    \caption{Schematic overview of the SPRINT framework.}
    \label{fig:placeholder}
\end{figure*}

\section{Related Work}
\textbf{Few-shot learning} aims to recognize novel classes from limited labeled examples by learning transferable representations. Prototypical Networks \citep{snell2017prototypical} learn embeddings where classes cluster around prototypes computed as mean embeddings, enabling nearest-neighbor classification in metric space. Matching Networks \citep{NIPS201690e13578} employ attention mechanisms over support sets, while Relation Networks \citep{hu2018relation} learn non-linear distance metrics. Meta-learning approaches like MAML \citep{finn2017model} learn good initializations for rapid gradient-based adaptation to novel tasks. Few-shot learning has expanded beyond vision to graphs (e.g., \citep{10.1145/3442381.3450112,xiong2018one}) and tabular data. While STUNT \cite{nam2023stunt} introduces tabular self-training and SSPNet \cite{11310333} integrates semi-supervised pseudo-labeling, they are limited to static, single-session settings. Our work differs by tackling the incremental learning scenario where classes arrive sequentially and catastrophic forgetting must be prevented.

\textbf{Class-incremental learning} (CIL) addresses catastrophic forgetting when learning new classes sequentially. Traditional approaches include: (1) \textbf{regularization-based methods} that constrain parameter updates through knowledge distillation \citep{li2017learning} or elastic weight consolidation \citep{kirkpatrick2017overcoming}, (2) \textbf{replay-based methods} that store exemplars from previous classes for rehearsal \citep{Rebuffi_2017_CVPR,castro2018end}, and (3) \textbf{architecture-based methods} that allocate separate parameters for each task \citep{rusu2016progressive,yoon2017lifelong}. iCaRL \citep{Rebuffi_2017_CVPR} pioneered exemplar-based CIL by combining nearest-mean-of-exemplars classification with knowledge distillation. LUCIR \citep{hou2019learning} addresses classifier bias through cosine normalization and margin-based separation. However, these methods assume abundant data per novel class, making them unsuitable for few-shot scenarios.

\textbf{Few-shot class-incremental learning} (FSCIL) combines the challenges of few-shot learning and class-incremental learning. TOPIC \citep{tao2020few} first formalized this problem for vision, proposing to freeze the feature extractor after base training and learn neural gas networks for each session. FACT \citep{9878986} employs forward-compatible training with self-supervised learning to prepare embeddings for future classes. However, all existing FSCIL methods: (1) operate exclusively on image data with spatial inductive biases, (2) assume only k-shot labeled examples per novel class, ignoring unlabeled data, (3) are unable to retain base data memory due to strict memory constraints.

\textbf{Semi-supervised learning} leverages unlabeled data to improve model performance. Pseudo-labeling \citep{lee2013pseudo} iteratively assigns labels to high-confidence predictions. Consistency regularization methods like $\Pi$-Model \citep{laine2016temporal}, Mean Teacher \citep{tarvainen2017mean}, and UDA \citep{xie2020unsupervised} enforce prediction invariance under perturbations. FixMatch \citep{sohn2020fixmatch} unifies these approaches by combining weak-to-strong augmentation consistency with confidence-based pseudo-labeling. To our knowledge, no prior work has investigated semi-supervised learning for FSCIL, despite unlabeled data being abundant in many real-world tabular data applications.

\textbf{Positioning of Our Work.} Our work is the first to combine semi-supervised learning with FSCIL in tabular data applications. Unlike vision-based FSCIL methods that must operate under memory-constrained settings, we leverage the operational reality where: (1) base class data memory can be retained because of compliance (e.g., logging network traffic for compliance reasons), operational needs, and negligible memory requirements, and (2) unlabeled data are continuously collected (e.g., most of the network traffic data are not labeled). We show that even with the base data access, standard FSCIL methods suffer significant forgetting (17.32\%), and that semi-supervised pseudo-labeling reduces this to 5.24\% by enriching novel class representations.

\section{Problem Formulation}

We formalize \textbf{Few-Shot Class-Incremental Learning (FSCIL)} for tabular domains where operational constraints necessitate robust retention of historical knowledge. Learning proceeds over $T+1$ sessions. In the base session ($t=0$), the model learns classes $\mathcal{C}^{(0)}$ from a large labeled dataset $\mathcal{S}^{(0)} = \{(x_i, y_i)\}_{i=1}^{N}$. In incremental sessions $t \in \{1, \ldots, T\}$, disjoint novel classes $\mathcal{C}^{(t)}$ are introduced via $k$-shot support sets $\mathcal{S}^{(t)} = \{(x_i^{(t)}, y_i^{(t)})\}$, where $|\{j: y_j^{(t)} = c\}| = k$ for all $c \in \mathcal{C}^{(t)}$. Additionally, a pool of unlabeled data (e.g., unlabeled network traffic data) $\mathcal{U}^{(t)} = \{x_j^{(t)}\}_{j=1}^{U_t}$ is continuously available.

\textbf{Problem Definition.} 
%MK-Done: M^0 is not defined. please check. Is it model?
Given base data stored in a memory buffer $\mathcal{M}^{(0)} \subseteq \mathcal{S}^{(0)}$, few-shot novel class samples $\mathcal{S}^{(t)}$ ($k$ per class), and unlabeled data $\mathcal{U}^{(t)}$, learn an embedding function $f_\theta: \mathbb{R}^D \to \mathbb{R}^M$ (where $D$ is the input feature dimension) that: (1) maintains high accuracy on base classes $\mathcal{C}^{(0)}$ across all sessions without catastrophic forgetting, and (2) learns novel classes $\mathcal{C}^{(t)}$ from $k$ labeled examples.

%MK-Done: M^0 seems to mean memory. Please clarify this above. 

%MK-Done: actually you cannot store the entire network data. it is too costly even for networkds. Maybe we can soften this. - Yes, actually we do not use full network data, we use 2000 samples per base class, but the difference is we retain them duirng incremental sessions, different than constraints  of vision benchmarks. 

\textbf{Memory and Storage Assumption.} Diverging from the fixed-buffer constraints of vision benchmarks, we retain the base dataset as memory $\mathcal{M}^{(0)} \subseteq \mathcal{S}^{(0)}$. While retaining large high-velocity streams is infeasible, maintaining a robust, representative archive of historical signatures (e.g., the base training corpus) is operationally viable due to the negligible storage footprint of tabular records relative to high-dimensional images (e.g., a 40-feature tabular record requires $\approx 160$ bytes, whereas a single $224{\times}224$ image requires $\approx 150$ KB, a $\approx 1000\times$ difference). Furthermore, in domains like Network Intrusion Detection, data retention is often required for audit compliance for certain time periods, and maintaining high detection rates on previously seen attack signatures is critical to avoid exploitable vulnerabilities. We empirically validate that this strategy scales effectively across diverse fields, ranging from healthcare to environmental monitoring.

\textbf{Memory and Storage Assumption.} Diverging from the fixed-buffer constraints of vision benchmarks, we retain the base dataset as memory $\mathcal{M}^{(0)} = \mathcal{S}^{(0)}$. While retaining large high-velocity streams is infeasible, maintaining a robust, representative archive of historical signatures (e.g., the base training corpus) is operationally viable due to the negligible storage footprint of tabular records relative to high-dimensional images. Furthermore, in domains like Network Intrusion Detection, data retention is often required for audit compliance for certain time period and maintaining high detection rates on previously seen attack signatures is critical to avoid exploitable vulnerabilities.  We empirically validate that this strategy scales effectively across diverse fields, ranging from healthcare to environmental monitoring.

%MK-Done: If we have never labeled a specific instance before and it is novel, how come we have unlabeled samples of it in the dataset ?? We must justify this assumption more clearly. This could be attacked easily. - We assume this because in streaming domains (like NIDS or Healthcare), there is a temporal gap between collection and annotation. When a new class appears (e.g., a new attack campaign), the system automatically collects thousands of packets immediately. These sit in the buffer as unlabeled data. It might take hours for an expert to identify and label just 5 of them ($k$-shot). Therefore, at the moment of learning, it is realistic to have a large pool of unlabeled instances of the new class alongside the few labeled ones.
\textbf{Unlabeled Data Composition.} To emulate realistic open-world streams, we model the \textbf{operational lag between data ingestion and expert annotation}. In high-velocity domains like NIDS, unlabeled traffic arrives continuously; when a novel class emerges (e.g., a zero-day attack), instances accumulate in the system's log buffer long before an analyst identifies and labels the first few signatures. Consequently, while the labeled support set $\mathcal{S}^{(t)}$ is sparse ($k$-shot), the unlabeled pool $\mathcal{U}$ naturally contains latent samples of this current novel class ($y \in \mathcal{C}^{(t)}$), mixed with historical base data ($y \in \mathcal{C}^{(0)}$) and unlabeled instances of future classes ($y \in \mathcal{C}^{(>t)}$). This creates a challenging pseudo-labeling scenario requiring the model to distinguish relevant signals from diverse noise.

% To emulate realistic open-world conditions, the unlabeled pool is a mix of known base classes ($y \in \mathcal{C}^{(0)}$), current novel classes ($y \in \mathcal{C}^{(t)}$), and future unseen classes ($y \in \mathcal{C}^{(>t)}$). This creates a challenging pseudo-labeling scenario requiring the model to distinguish relevant samples from historical data and future outliers.

% We sample batches $\mathcal{U}_{\text{batch}}^{(t)} \subset \mathcal{U}$ to enrich $\mathcal{S}^{(t)}$ via semi-supervised learning.
% We employ episodic training with Prototypical Networks \citep{snell2017prototypical} for base session training. Each training episode samples $N_{\text{way}}$ classes from $\mathcal{C}^{(0)}$ and constructs support set $\mathcal{S}_{\text{ep}}$ (with $k_{\text{train}}$ samples per class) and query set $\mathcal{Q}_{\text{ep}}$.

% \textbf{Prototype Computation.} For each class $c$ in the episode:
% \begin{equation}
% \mathbf{p}_c^{(0)} = \frac{1}{|\mathcal{S}_{c,\text{ep}}|} \sum_{(x_i, y_i) \in \mathcal{S}_{c,\text{ep}}} f_{\theta}(x_i)
% \end{equation}

% \textbf{Classification via Negative Distance Softmax:}
% \begin{equation}
% p_\theta(y = c \mid x_q) = \frac{\exp(-\|f_\theta(x_q) - \mathbf{p}_c^{(0)}\|_2^2)}{\sum_{c' \in \mathcal{C}_{\text{ep}}} \exp(-\|f_\theta(x_q) - \mathbf{p}_{c'}^{(0)}\|_2^2)}
% \end{equation}

% \textbf{Base Training Loss:}
% \begin{equation}
% \mathcal{L}_{\text{base}} = - \frac{1}{|\mathcal{Q}_{\text{ep}}|} \sum_{(x_q, y_q) \in \mathcal{Q}_{\text{ep}}} \log p_\theta(y = y_q \mid x_q)
% \end{equation}
\section{Base Session Training}
We employ episodic training with Prototypical Networks \citep{snell2017prototypical} for base session training. Each training episode samples $N_{\text{way}}$ classes from $\mathcal{C}^{(0)}$ and constructs a support set $\mathcal{S}_{\text{ep}}$ (with $k_{\text{train}}$ samples per class) and a query set $\mathcal{Q}_{\text{ep}}$. For each class $c$ in the episode, the prototype is computed as $\mathbf{p}_c^{(0)} = \frac{1}{|\mathcal{S}_{c,\text{ep}}|} \sum_{(x_i, y_i) \in \mathcal{S}_{c,\text{ep}}} f_{\theta}(x_i)$. Classification is performed via a negative distance softmax, $p_\theta(y = c \mid x_q) = \frac{\exp(-\|f_\theta(x_q) - \mathbf{p}_c^{(0)}\|_2^2)}{\sum_{c' \in \mathcal{C}_{\text{ep}}} \exp(-\|f_\theta(x_q) - \mathbf{p}_{c'}^{(0)}\|_2^2)}$, and the base training loss is $\mathcal{L}_{\text{base}} = - \frac{1}{|\mathcal{Q}_{\text{ep}}|} \sum_{(x_q, y_q) \in \mathcal{Q}_{\text{ep}}} \log p_\theta(y = y_q \mid x_q)$.

\section{Incremental Session Training}

\subsection{Encoder Update with Forgetting Mitigation}

\textbf{Encoder Parameter Update.} Unlike methods that freeze the encoder after base training, we continue updating $\theta$ during incremental sessions. This allows the embedding space to adapt to novel classes while maintaining discriminability for base classes. At each incremental session, training loss is composed of two complementary losses:
\begin{enumerate}
    \item \textbf{Base class loss} $\mathcal{L}_{\text{proto}}^{(t)}$ generated on data sampled from $\mathcal{M}^{(0)}$ (as defined in Eq. \ref{eq:proto_loss}).
    \item \textbf{Novel class semi-supervised loss} $\mathcal{L}_{\text{semi}}^{(t)}$  combining labeled $\mathcal{S}^{(t)}$ and pseudo-labeled $\mathcal{U}^{*(t)}$  (Eq. \ref{eq:proto_semi_loss}).
\end{enumerate}

This mixed episodic training ensures $\theta$ simultaneously optimizes for both base class retention and novel class learning. Critically, we do \textbf{not} employ explicit regularization penalties (e.g., knowledge distillation). Instead, forgetting prevention emerges from joint optimization: gradients from $\mathcal{L}_{\text{proto}}^{(t)}$ continuously replay base class discrimination tasks, maintaining the embedding space's ability to separate base classes.

\subsection{Confidence-Based Pseudo-Labeling}
At the onset of an incremental session $t$, we first initialize a prototype for each new class $c \in \mathcal{C}^{(t)}$ using its limited $k$-shot labeled examples: $\mathbf{p}_c^{(t,0)} = \frac{1}{k} \sum_{(x_i, y_i) \in \mathcal{S}^{(t)}, y_i = c} f_{\theta}(x_i)$. We then project the entire unlabeled pool $\mathcal{U}$ into the embedding space and compute the Euclidean distance $d_{j,c}$ between each unlabeled sample $x_j$ and all currently active prototypes: $
d_{j,c} = \|f_{\theta}(x_j) - \mathbf{p}_c\|_2^2, \quad \forall c \in \mathcal{C}^{(\leq t)}
$.

Each sample is assigned a pseudo-label $\hat{y}_j = \arg\min_{c} d_{j,c}$ corresponding to its nearest prototype. To mitigate the risk of noisy predictions, we filter these assignments by selecting only the most confident samples. Specifically, for each novel class $c \in \mathcal{C}^{(t)}$, we retain the top-$m$ samples with the smallest distances to the class prototype, forming the high-confidence set:
\begin{equation}
\mathcal{U}_c^{*(t)} = \{x_j \in \mathcal{U} : \hat{y}_j = c \text{ and } d_{j,c} \text{ ranks in top-}m\}
\end{equation}
These filtered subsets are aggregated into a final pseudo-labeled pool $\mathcal{U}^{*(t)} = \bigcup_{c \in \mathcal{C}^{(t)}} \mathcal{U}_c^{*(t)}$, which is subsequently used to augment the training distribution in the semi-supervised phase.

\subsection{Episodic Training with Mixed Supervision}

Our training strategy interleaves base class rehearsal with semi-supervised novel class learning within each episode. Unlike methods that alternate between different episode types across training iterations, we compute \textbf{both} losses in every episode and optimize their weighted combination.

\textbf{Sub-episode 1: Base Class Rehearsal.} We first reinforce base knowledge by randomly sampling $N_{\text{way}}$ classes from $\mathcal{C}^{(0)}$. For each selected class $c$, we draw $k$ support and $q$ query samples from the memory buffer $\mathcal{M}^{(0)}$ to construct the base support set $\mathcal{S}_{\text{base}}$ (size $N_{\text{way}} \cdot k$) and query set $\mathcal{Q}_{\text{base}}$ (size $N_{\text{way}} \cdot q$). We compute base class prototypes $\mathbf{p}_c^{\text{base}}$ by averaging the embeddings of support samples.

% \begin{equation}
% \mathbf{p}_c^{\text{base}} = \frac{1}{k} \sum_{(x_i, y_i) \in \mathcal{S}_{\text{base}}, y_i = c} f_\theta(x_i), \quad \forall c \in \mathcal{C}_{\text{ep}}
% \end{equation}

Subsequently, we compute the base class prototypical loss $\mathcal{L}_{\text{proto}}^{(t)}$ by minimizing the negative log-probability of the true class for the query samples:
\begin{equation}
\label{eq:proto_loss}
\mathcal{L}_{\text{proto}}^{(t)} = - \frac{1}{|\mathcal{Q}_{\text{base}}|} \sum_{(x_q, y_q) \in \mathcal{Q}_{\text{base}}} \log p_\theta(y = y_q \mid x_q)
\end{equation}
where the probability is given by the softmax over negative Euclidean distances:
\begin{equation}
p_\theta(y = c \mid x_q) = \frac{\exp(-\|f_\theta(x_q) - \mathbf{p}_c^{\text{base}}\|_2^2)}{\sum_{c' \in \mathcal{C}_{\text{ep}}} \exp(-\|f_\theta(x_q) - \mathbf{p}_{c'}^{\text{base}}\|_2^2)}
\end{equation}

\textbf{Sub-episode 2: Semi-Supervised Novel Class Learning.} To enrich the representation of novel classes, we construct a combined data pool $\mathcal{D}_{\text{combined}} = \bigcup_{c \in \mathcal{C}^{(t)}} (\mathcal{S}^{(t)}_c \cup \mathcal{U}_c^{*(t)})$ for each class $c \in \mathcal{C}^{(t)}$, aggregating the $k$ labeled examples with the high-confidence pseudo-labeled samples selected in Phase 1. From this pool, we form the novel support set $\mathcal{S}_{\text{semi}}$ and query set $\mathcal{Q}_{\text{semi}}$ by sampling $k$ and $q$ instances per class, respectively; if the pseudo-labeled count is insufficient, we oversample labeled data with replacement. Crucially, to enforce decision boundaries between established and emerging classes, we augment these sets with samples from $N_{\text{way}}$ base classes drawn from $\mathcal{M}^{(0)}$. This yields the full semi-supervised support set $\mathcal{S}_{\text{semi}}^{\text{full}} = \mathcal{S}_{\text{base}} \cup \mathcal{S}_{\text{semi}}$ and query set $\mathcal{Q}_{\text{semi}}^{\text{full}} = \mathcal{Q}_{\text{base}} \cup \mathcal{Q}_{\text{semi}}$, containing a total of $(N_{\text{way}} + |\mathcal{C}^{(t)}|) \cdot k$ support samples. Prototypes $\mathbf{p}_c^{\text{semi}}$ are computed over this mixed support set: $
\mathbf{p}_c^{\text{semi}} = \frac{1}{|\mathcal{S}_{c,\text{semi}}^{\text{full}}|} \sum_{(x_i, \hat{y}_i) \in \mathcal{S}_{\text{semi}}^{\text{full}}, \hat{y}_i = c} f_\theta(x_i)
$ and the semi-supervised loss is derived from the mixed query set:
\begin{equation}
\label{eq:proto_semi_loss}
\mathcal{L}_{\text{semi}}^{(t)} = - \frac{1}{|\mathcal{Q}_{\text{semi}}^{\text{full}}|} \sum_{(x_q, \hat{y}_q) \in \mathcal{Q}_{\text{semi}}^{\text{full}}} \log p_\theta(y = \hat{y}_q \mid x_q)
\end{equation}
\textbf{Joint Optimization.} Ideally, the model should retain base knowledge while adapting to novel clusters. We achieve this by optimizing a weighted combination of the base rehearsal loss and the semi-supervised loss at each episode $e$, controlled by a balancing coefficient $\beta \in [0, 1]$:
\begin{equation}
\mathcal{L}^{(t)}(e) = \beta \cdot \mathcal{L}_{\text{proto}}^{(t)} + (1-\beta) \cdot \mathcal{L}_{\text{semi}}^{(t)}
\end{equation}
Parameters are updated via standard gradient descent $\theta \leftarrow \theta - \eta \nabla_\theta \mathcal{L}^{(t)}(e)$.

\textbf{Implicit Forgetting Prevention.} Distinct from methods that rely on explicit regularization penalties (e.g., Elastic Weight Consolidation or distillation), our approach enforces retention implicitly through the loss structure. The $\mathcal{L}_{\text{proto}}^{(t)}$ term continuously replays base discrimination tasks, ensuring the embedding space maintains separability for $\mathcal{C}^{(0)}$. Simultaneously, $\mathcal{L}_{\text{semi}}^{(t)}$ aligns novel class embeddings relative to these stable base anchors, utilizing the abundant pseudo-labeled data to form tighter, more robust clusters than would be possible with only $k$ labeled examples.

\section{Computational Complexity Analysis}
% We analyze the computational efficiency of SPRINT, showing that it maintains inference parity with standard ProtoNet while incurring moderate training overhead. Throughout this analysis, $D$ denotes the input feature dimensionality, $M$ denotes the embedding dimensionality, and $|\mathcal{C}^{(\leq t)}|$ denotes the total number of classes seen up to session $t$.
\textbf{Inference Complexity.} At inference time, SPRINT maintains\textbf{ identical computational complexity to standard Prototypical Networks} \citep{snell2017prototypical}, as pseudo-labeling only occurs during incremental training. The semi-supervised component introduces \textbf{zero inference overhead}, making SPRINT suitable for production deployment with real-time latency requirements.

% \textbf{Inference Complexity}. At inference time, SPRINT maintains\textbf{ identical computational complexity to standard Prototypical Networks}. For a test sample $x_{\text{test}} \in \mathbb{R}^D$, prediction requires computing the embedding $f_\theta(x_{\text{test}}) \in \mathbb{R}^M$ with cost $O(D \cdot M)$, followed by distance computation $\|f_\theta(x_{\text{test}}) - \mathbf{p}_c\|_2^2$ to all $|\mathcal{C}^{(\leq t)}|$ prototypes with cost $O(|\mathcal{C}^{(\leq t)}| \cdot M)$, and finally classification via argmin with cost $O(|\mathcal{C}^{(\leq t)}|)$. The total inference complexity is therefore $
% O(D \cdot M + |\mathcal{C}^{(\leq t)}| \cdot M)$.

% This matches standard ProtoNet exactly, as pseudo-labeling only occurs during training. The semi-supervised component introduces \textbf{zero inference overhead}, making SPRINT suitable for production deployment with real-time latency requirements.

\textbf{Training Efficiency and Scalability}. A critical advantage of SPRINT is its \textbf{operational efficiency}. Unlike replay-based distillation methods (e.g., iCaRL \cite{Rebuffi_2017_CVPR}) where computational cost scales linearly with the size of the memory buffer ($O(|\mathcal{M}^{(0)}|)$), SPRINT employs sparse episodic sampling (i.e., constructing episodes by sampling small support and query sets from the memory buffer). This mechanism decouples training latency from memory capacity, allowing the system to leverage the memory buffer for stability without incurring the computational penalty of dense rehearsal. Consequently, SPRINT achieves an order-of-magnitude reduction in incremental adaptation time compared to dense replay baselines (empirically $\approx 18\times$ faster under identical memory constraints), while avoiding the prohibitive second-order costs of optimization-based meta-learning. Detailed Big-O derivations and empirical runtime comparisons are provided in Appendix~\ref{sec:complexity}.

\begin{table*}[h!]
\centering
\caption{Detailed configuration of dataset splits, class partitions, and base class memory budgets ($M^{(0)}$ samples per base class) used in the FSCIL experiments. The \textbf{Incr. Classes} column lists categories introduced sequentially in sessions $1 \dots S$.}
\label{tab:dataset_splits}
\resizebox{\textwidth}{!}{%
\begin{tabular}{l c p{5.5cm} p{5.5cm} c}
\toprule
\textbf{Dataset} & \textbf{Feat.} & \textbf{Base Classes ($C_{base}$)} & \textbf{Incr. Classes ($C_{novel}$)} & \textbf{$M^{(0)}$ / Class} \\
\midrule
\textbf{ACI-IoT-2023} \cite{qacj-3x32-23} & 79 & Benign, Dictionary, OS Scan, Port Scan, Ping Sweep, Vuln. Scan & ICMP Flood, Slowloris, SYN Flood, DNS Flood & 2,000 \\
\midrule
\textbf{CIC-IDS2017} \cite{sharafaldin2018toward} & 77 & BENIGN, PortScan, DDoS, FTP-Patator, SSH-Patator & DoS Hulk, DoS GoldenEye, DoS Slowhttptest, DoS slowloris & 2,000 \\
\midrule
\textbf{CIC-IoT2023} \cite{neto2023ciciot2023} & 39 & BENIGN, DNS Spoofing, DoS: HTTP, SYN, TCP, UDP, Mirai: Greeth, Greip, UDPPlain, MITM, Recon: Host, OS, Port, Vuln. Scan & DDoS: Ack, HTTP, ICMP, ICMP-FRAG, PSHACK, RSTFIN, SynonymousIP, SYN, TCP, UDP, UDP-Frag & 2,000 \\
\midrule
\textbf{Obesity} \cite{Palechor2019DatasetFE} & 31 & Insufficient Wgt, Normal Wgt, Overweight Lvl I, Overweight Lvl II & Obesity Type I, Obesity Type II, Obesity Type III & 100 \\
\midrule
\textbf{CovType} \cite{covertype_31} & 54 & Spruce/Fir, Lodgepole Pine, Ponderosa Pine, Cottonwood/Willow & Aspen, Douglas-fir, Krummholz & 2,000 \\
\midrule
\textbf{MNIST} \cite{lecun2010mnist} & 784 & Digits: 0, 1, 2, 3, 4, 5 & Digits: 6, 7, 8, 9 & 2,000 \\
\bottomrule
\end{tabular}%
}
\end{table*}

% We analyze the computational complexity of SPRINT, accounting for both the one-time pseudo-labeling phase and the per-episode mixed supervision cost. Throughout this analysis, $k$ denotes support samples per class, and $q$ denotes query samples per class.

\section{Experiments}

% In this section, we present a comprehensive empirical evaluation of SPRINT against existing state-of-the-art FSCIL frameworks. To assess cross-domain generalization, we benchmark our approach across a diverse suite of datasets, including healthcare diagnostics, network intrusion detection, and environmental monitoring. Our ablation studies validate the synergistic effect of semi-supervised learning within the few-shot regime.

% , demonstrating significant performance gains under data scarcity. Finally, we provide a granular visualization of SPRINT’s incremental trajectory, illustrating its ability to seamlessly integrate novel classes while preventing catastrophic forgetting of old classes.

\begin{table*}[t]
\centering
\caption{Summary of results on 6 datasets (5-shot setting). PD: Performance Dropping rate ($A_0 - A_{last}$), Acc: Last session accuracy ($A_{last}$). Avg-PD and Avg-Acc represent the average values across all datasets. Best results are in \textbf{bold}, second best in \textit{italics}.}
\begin{small}
\begin{sc}
\resizebox{\textwidth}{!}{
\begin{tabular}{l|cc|cc|cc|cc|cc|cc|cc}
\hline
\multirow{2}{*}{Method} & \multicolumn{2}{c|}{ACI-IoT-2023} & \multicolumn{2}{c|}{Obesity} & \multicolumn{2}{c|}{CICIDS2017} & \multicolumn{2}{c|}{CIC-IoT-2023} & \multicolumn{2}{c|}{CovType} & \multicolumn{2}{c|}{MNIST} & \multicolumn{2}{c}{\textbf{Average}} \\
  & PD $\downarrow$ & Acc $\uparrow$ & PD $\downarrow$ & Acc  $\uparrow$ & PD $\downarrow$ & Acc  $\uparrow$ & PD $\downarrow$ & Acc  $\uparrow$ & PD $\downarrow$ & Acc  $\uparrow$ & PD $\downarrow$ & Acc  $\uparrow$ &\textbf{ Avg-PD $\downarrow$} & \textbf{Avg-Acc  $\uparrow$} \\
\hline
CNN ProtoNet & 9.49 & 76.22 & \textit{5.04} & 55.53 & 23.51 & 69.91 & 0.99 & 53.56 & \textbf{9.72} & 45.87 & \textit{8.89} & \textit{81.10} & \textit{9.61} & 63.70 \\
FACT & \textit{8.06} & 69.93 & 10.71 & 14.29 & \textit{15.17} & 18.42 & \textit{-2.46} & 53.68 & 13.47 & 16.84 & 25.00 & 40.65 & 11.66 & 35.63 \\
MAML & 33.92 & 47.27 & 23.95 & 37.28 & 43.96 & 42.67 & 18.22 & 35.39 & 32.62 & 38.28 & 45.59 & 48.42 & 33.04 & 41.55 \\
Neuron Expansion & 12.33 & 85.79 & 22.60 & 59.45 & 20.92 & 74.01 & 30.71 & 60.39 & 30.33 & 50.08 & 27.13 & 70.90 & 24.00 & 66.77 \\
ProtoNet & 12.03 & 84.16 & 10.71 & 71.67 & 27.05 & 69.21 & 2.02 & 59.85 & 11.44 & 54.91 & 23.87 & 70.97 & 14.52 & 68.46 \\

semi-Super-ProtoNet & 11.04 & 85.13 & 12.25 & 70.11 & 28.79 & 67.47 & 1.02 & 60.84 & \textit{10.47} & \textit{55.63} & 23.87 & 70.94 & 14.57 & 68.35 \\
iCaRL & 9.81 & \textit{89.18} & 9.70 & \textit{74.80} & 17.78 & \textit{80.23} & 6.27 & \textbf{68.86} & 33.43 & 53.47 & 26.95 & 71.00 & 17.32 & \textit{72.92} \\
\hline
SPRINT & \textbf{2.54} & \textbf{93.63} & \textbf{3.61} & \textbf{78.76} & \textbf{13.45} & \textbf{82.80} & \textbf{-3.94} & \textit{65.81} & 10.62 & \textbf{58.36} & \textbf{5.18} & \textbf{84.85} & \textbf{5.24} & \textbf{77.37} \\

\hline
\end{tabular}
}
\end{sc}
\end{small}
\label{tab:summary_results_with_avg}
\end{table*}

\begin{figure*}[t]
    \centering
    % First Subfigure
    \begin{subfigure}[b]{0.49\textwidth}
        \centering
        \includegraphics[width=\textwidth]{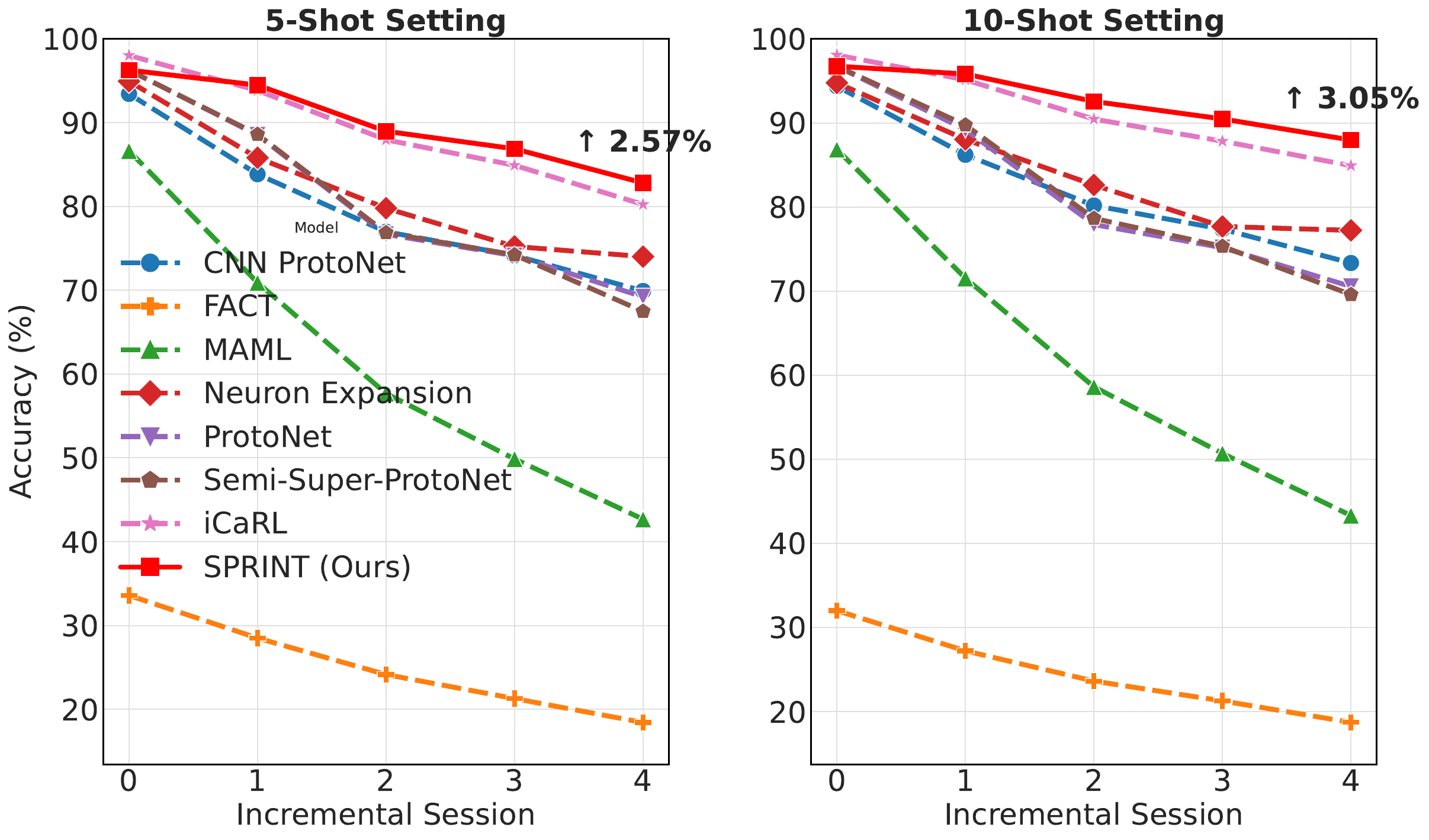}
        \caption{CIC-IDS2017}
        \label{fig:first}
    \end{subfigure}
    \hfill % Adds flexible space between images
    % Second Subfigure
    \begin{subfigure}[b]{0.49\textwidth}
        \centering
        \includegraphics[width=\textwidth]{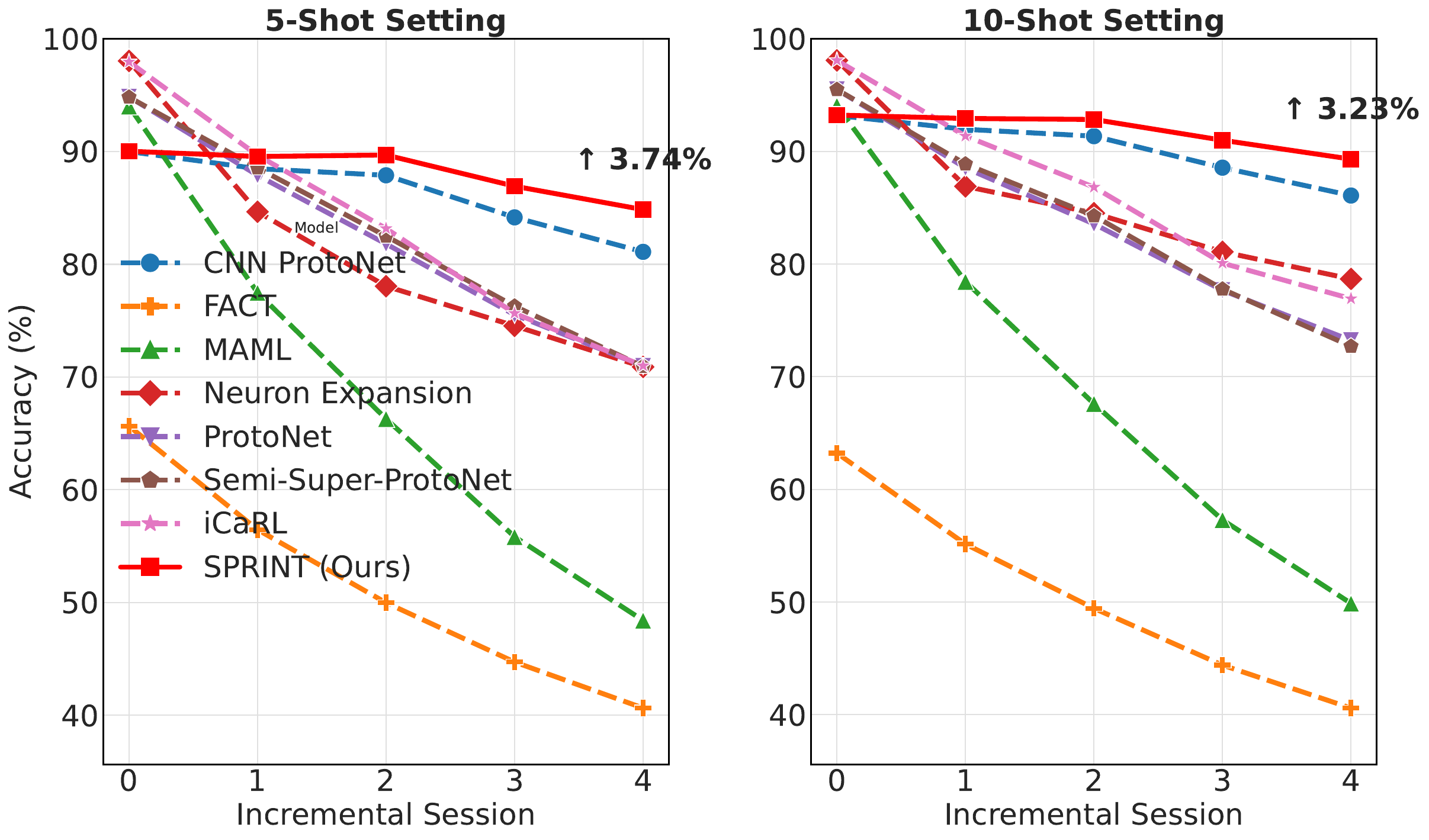}
        \caption{MNIST}
        \label{fig:second}
    \end{subfigure}
    
    \caption{Session-wise accuracy comparison on the a) CICIDS2017 and b) MNIST datasets. The annotation highlights the performance margin of SPRINT over the second-best method in the final incremental session.}
    \label{fig:model_comparison_sessions}
\end{figure*}

\begin{figure}[t]
    \centering
    \includegraphics[width=\linewidth]{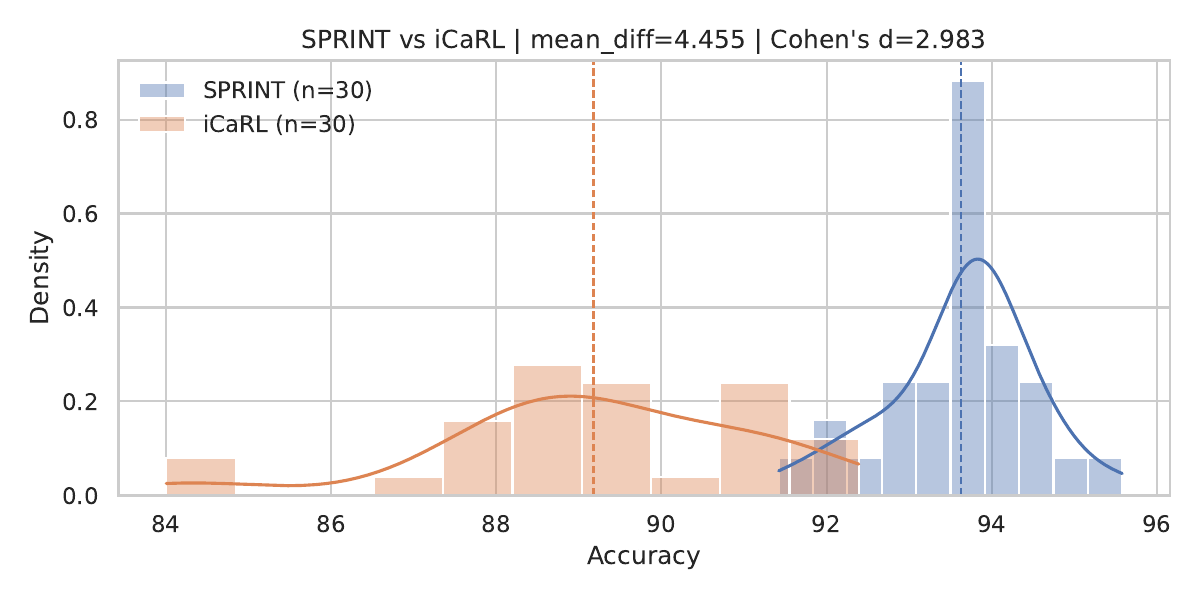}
    \caption{\textbf{Distribution of Top-1 Accuracy (Final Session).} Kernel Density Estimate (KDE) and histogram comparison of SPRINT vs. iCaRL over 30 independent runs on the ACI-IoT-2023 dataset. SPRINT exhibits a sharp, high-confidence peak around 93.6\%, demonstrating superior stability and minimal variance compared to the broader distribution of iCaRL.}
    \label{fig:statistical_sig}
\end{figure}

\begin{figure}[t]
    \centering
    % --- TOP ROW ---
    \begin{subfigure}[b]{0.48\columnwidth}
        \centering
        \includegraphics[width=\textwidth]{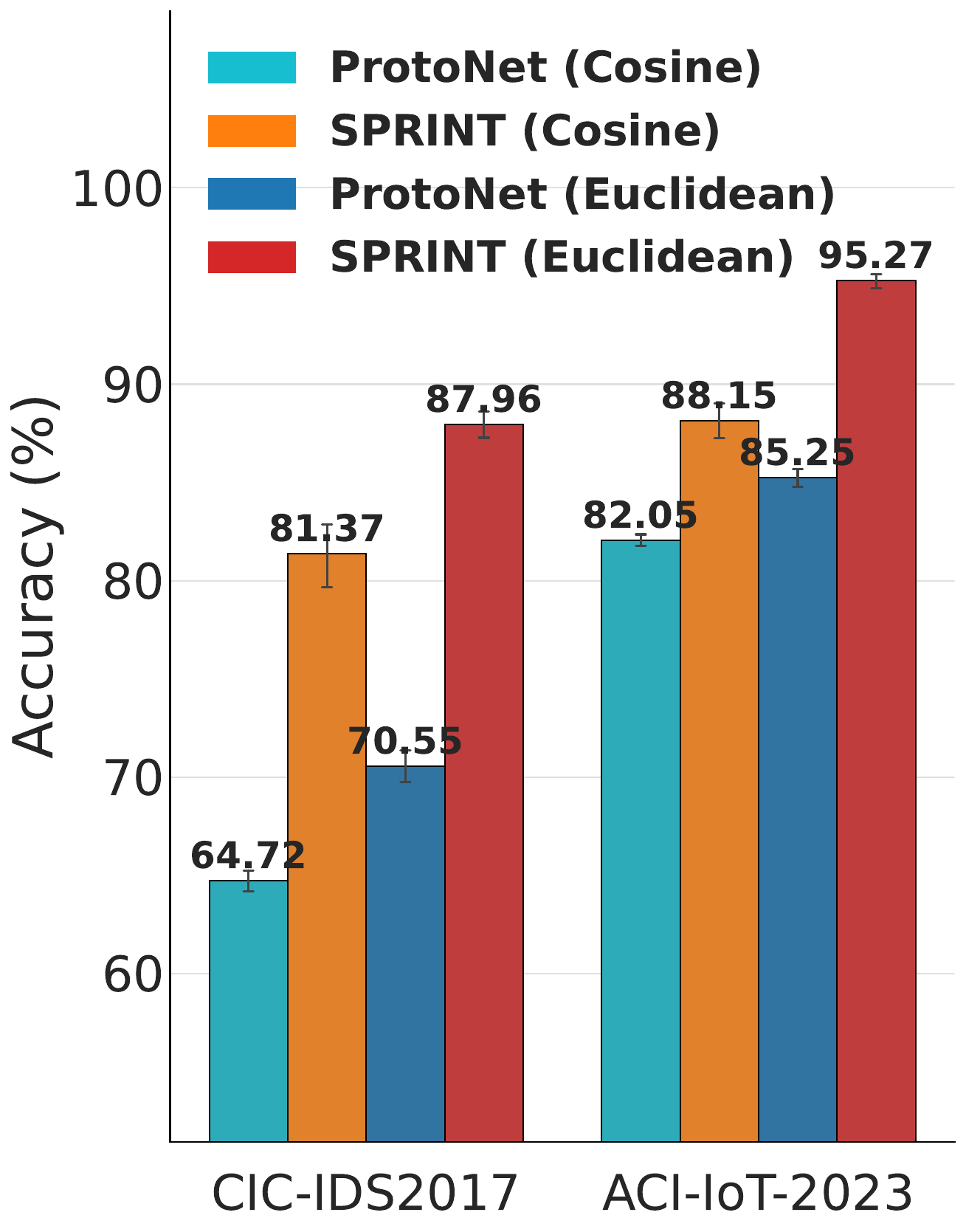}
        \caption{\textbf{Distance Metric Analysis.}}
        \label{fig:distance_comparison}
    \end{subfigure}%
    \hfill 
    \begin{subfigure}[b]{0.48\columnwidth}
        \centering
        \includegraphics[width=\textwidth]{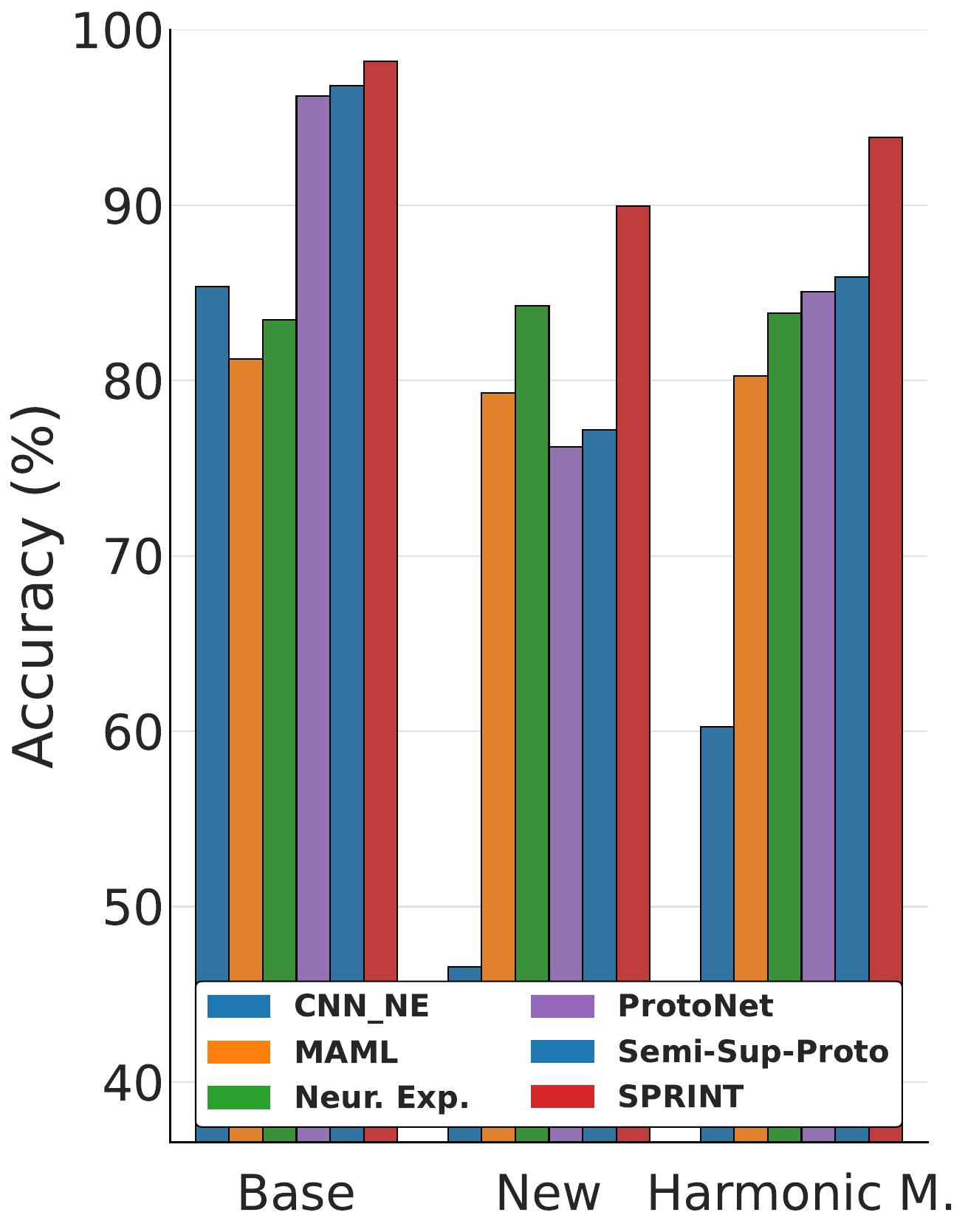}
        \caption{\textbf{Non-incremental Setting.}}
        \label{fig:nway_kshot}
    \end{subfigure}

    \vspace{1em} % Adds some vertical space between the rows

    % --- BOTTOM ROW ---
    \begin{subfigure}[b]{0.48\columnwidth}
        \centering
        \includegraphics[width=\textwidth]{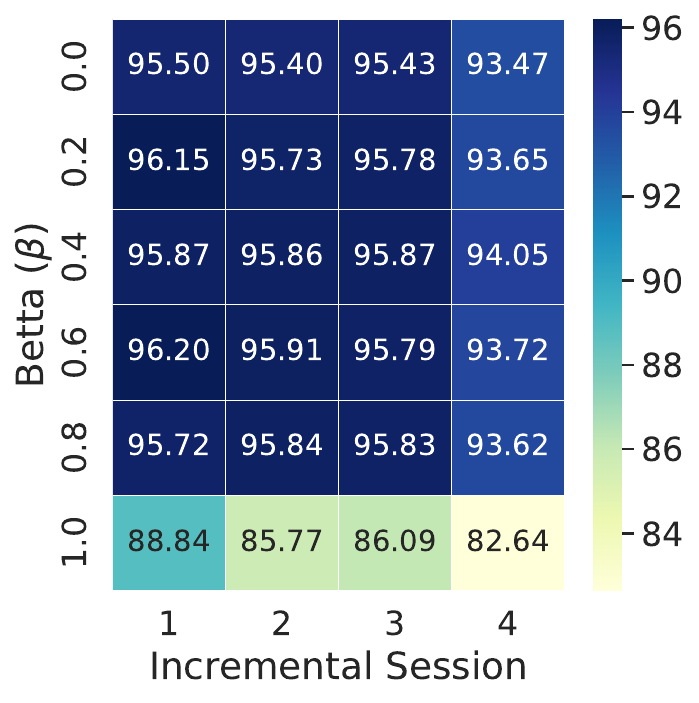}
        \caption{\textbf{$\beta$ Sensitivity.} }
        \label{fig:betta_ablations}
    \end{subfigure}%
    \hfill 
    \begin{subfigure}[b]{0.48\columnwidth}
        \centering
        \includegraphics[width=\textwidth]{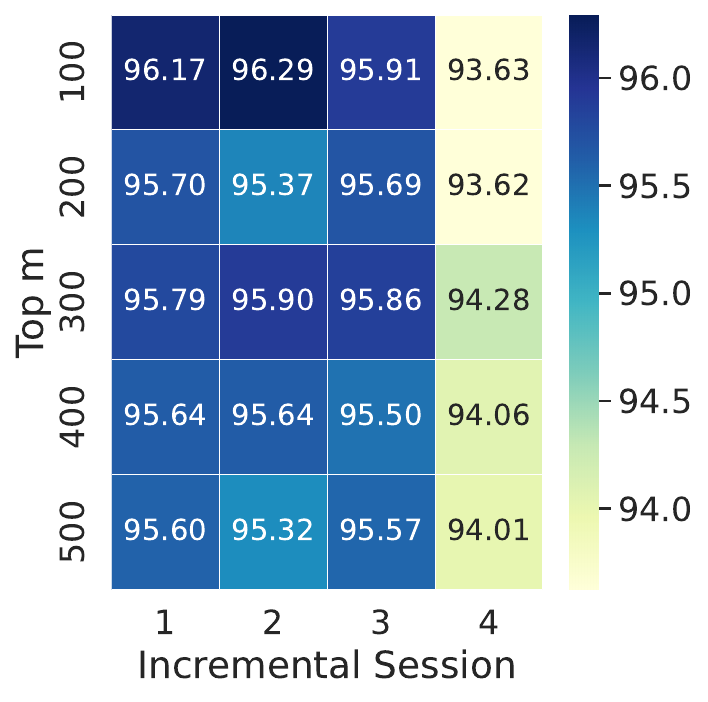}
        \caption{\textbf{Pseudo-label Pool Size}}
        \label{fig:number_of_pseudo_samples}
    \end{subfigure}
    
    \caption{\textbf{Design Analysis and Ablation Studies.} (a) Impact of distance metrics on classification accuracy. 
     %MK-Done: hard to see which one is maml which one is sprint. Please use something else for Maml in figure 4-b.
    (b) Evaluation in non-incremental setting. (c) Sensitivity of the loss balancing term $\beta$. (d) Influence of the number of pseudo-samples on Top-1 accuracy.}
    \label{fig:four_grid}
\end{figure}

\begin{figure}[t]
    \centering
    % --- TOP ROW ---
    \begin{subfigure}[b]{\columnwidth}
        \centering
        \includegraphics[width=\textwidth]{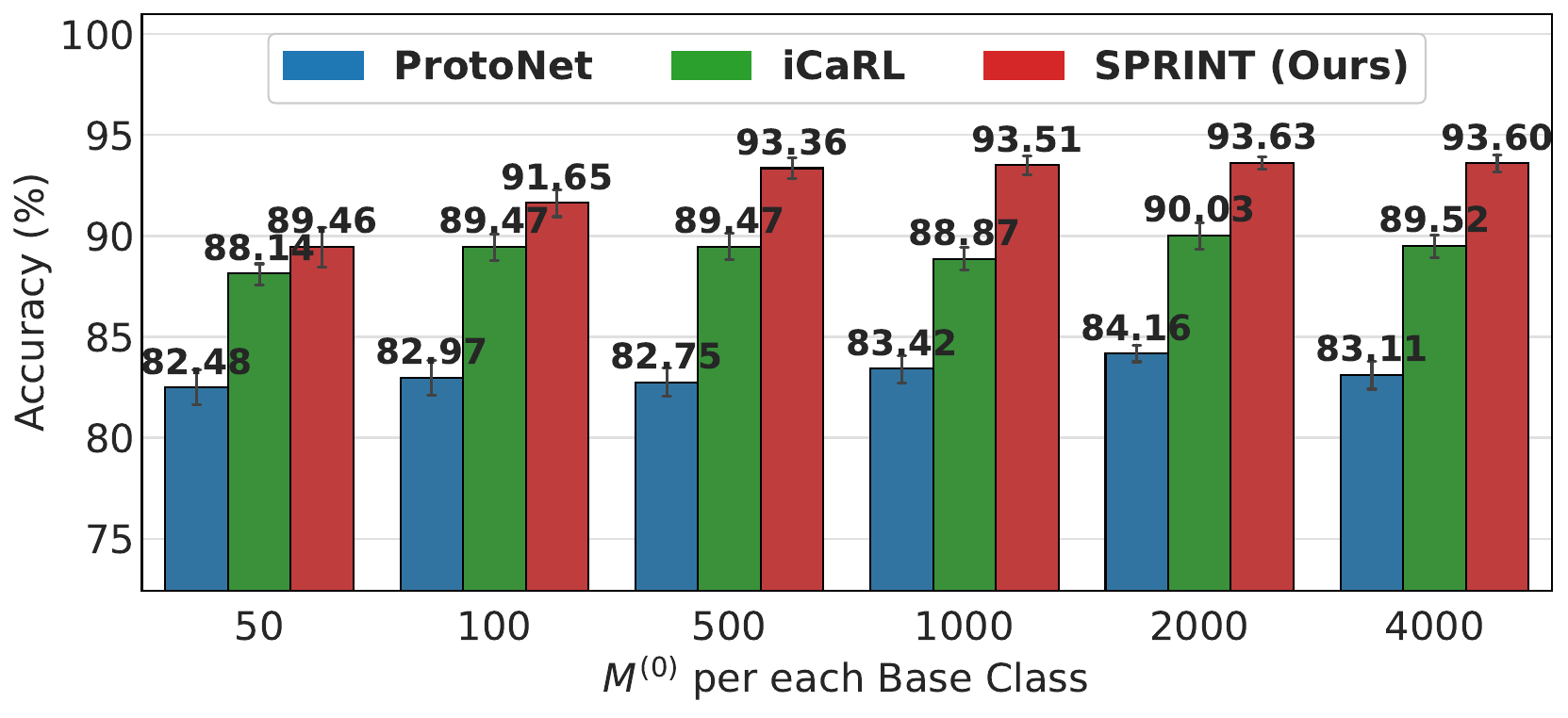}
        \caption{\textbf{Effect of Base Class Memory Budget ($M^{(0)}$).} Sensitivity analysis of Top-1 accuracy on the final session of ACI-IoT-2023.}
        \label{fig:memory_budget}
    \end{subfigure}%

    \vspace{1em} % Adds some vertical space between the rows

    % --- BOTTOM ROW ---
    \begin{subfigure}[b]{0.48\columnwidth}
        \centering
        \includegraphics[width=\textwidth]{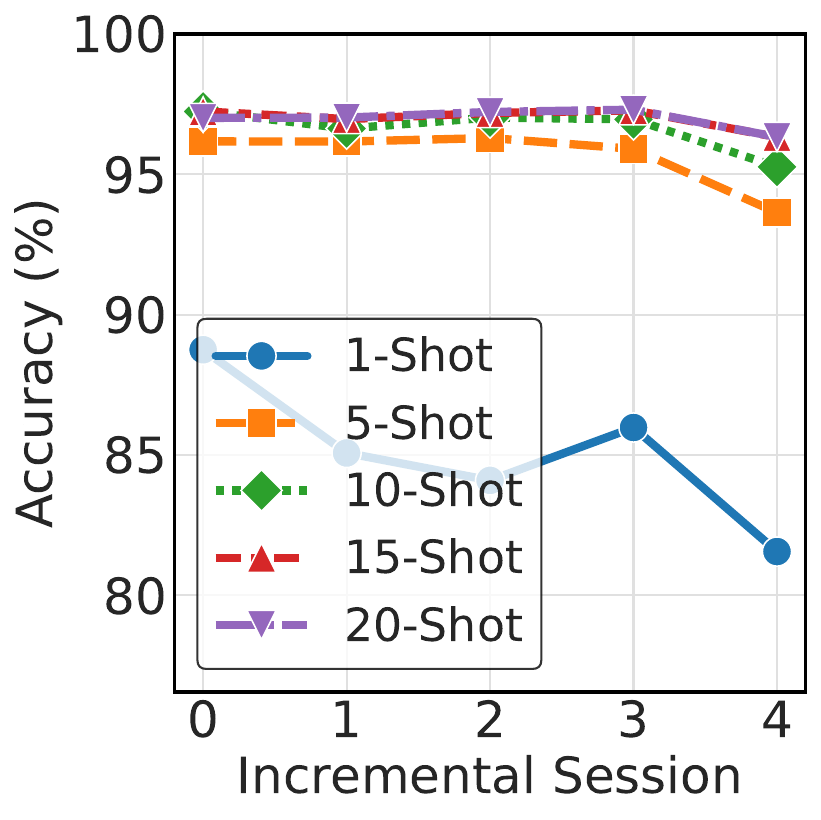}
        \caption{\textbf{Shot Availability ($k$).}}
        \label{fig:shots}
    \end{subfigure}%
    \hfill 
    \begin{subfigure}[b]{0.48\columnwidth}
        \centering
        \includegraphics[width=\textwidth]{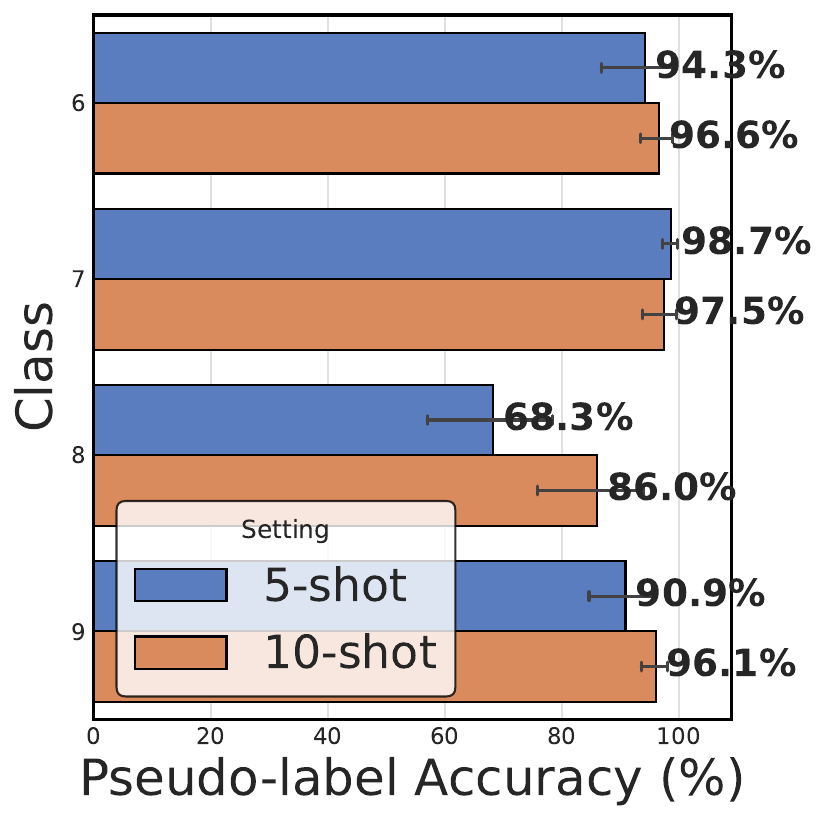}
        \caption{\textbf{Pseudo-label Reliability.}}
        \label{fig:pseudo-label-accuracy}
    \end{subfigure}
    
    \caption{\textbf{Sensitivity and Component Analysis.} (a) Influence of the memory budget $M^{(0)}$ on final retention. (b) model accuracy vs shot number. (c) Breakdown of pseudo-label accuracy, demonstrating the effectiveness of the semi-supervised component.}
    \label{fig:three_grid_analysis}
\end{figure}

%\subsection{Experimental Setup}
\textbf{Datasets and Experimental Protocols.} To evaluate cross-domain robustness, we utilize a diverse suite of benchmarks spanning cybersecurity (ACI-IoT-2023 \cite{qacj-3x32-23}, CIC-IDS2017 \cite{sharafaldin2018toward}, CIC-IoT2023 \cite{neto2023ciciot2023}), healthcare (Obesity \cite{Palechor2019DatasetFE}), ecological classification (CovType \cite{covertype_31}), and pattern recognition (MNIST \cite{lecun2010mnist}). \textbf{These datasets characterize a wide spectrum of feature space properties, ranging from high-dimensional inputs (up to 784 features) to compact feature sets (as low as 31 features).} Our evaluation follows a rigorous incremental schedule where each session introduces one novel class under data scarcity ($k$-shot labeled instances, where $k \in \{5, 10\}$). Regarding base class retention, although the source datasets offer massive labeled pools (spanning $10^4$ to $10^6$ instances), we strictly limit our base memory $\mathcal{M}^{(0)}$ to 2,000 samples per class for all large-scale tables (except Obesity, which is naturally smaller). This constraint ensures our evaluation remains rigorous and comparable to replay-based CIL standards \citep{Rebuffi_2017_CVPR}, preventing the trivialization of the forgetting challenge while reflecting the storage efficiency of tabular features. We also perform strict memory ablations (detailed in Section \ref{sec:robustness}) to benchmark against even tighter constraints. A summary of dataset statistics and experimental splits is presented in Table \ref{tab:dataset_splits}, while detailed descriptions of the class partitions are provided in Appendix~\ref{sec:data_splits}.

% \textbf{Datasets.} To evaluate the cross-domain robustness of SPRINT, we utilize a diverse suite of benchmarks spanning cybersecurity (ACI-IoT-2023 \cite{qacj-3x32-23}, CIC-IDS2017 \cite{sharafaldin2018toward}, CIC-IoT2023 \cite{neto2023ciciot2023}), healthcare (Obesity \cite{Palechor2019DatasetFE}), ecological classification (CovType \cite{covertype_31}), and pattern recognition (MNIST \cite{lecun2010mnist}).  \textbf{Beyond domain diversity, these datasets characterize a wide spectrum of feature space properties, ranging from high-dimensional inputs (up to 784 features) to compact feature sets (as low as 31 features).}  Our evaluation follows a rigorous incremental schedule where each session introduces one novel class under extreme data scarcity ($k$-shot labeled instances, where $k \in \{5, 10\}$). A summary of dataset statistics and experimental splits is presented in Table \ref{tab:dataset_splits}, while detailed descriptions of the class partitions and justifications are provided in Appendix~\ref{sec:data_splits}.

\textbf{Implementation Details.} Our feature extractor $f_\theta$ consists of a 3-layer Multi-Layer Perceptron (MLP) with a hidden dimension of 1024 and ReLU nonlinearities. We employ the Adam optimizer with a learning rate of $\eta = 0.001$ and no weight decay. For a fair comparison, all competing models utilize this MLP backbone, with the exception of \text{CNN-ProtoNet}. The latter incorporates a CNN-based feature extractor, previously validated in incremental few-shot regimes \cite{10720176}, which we have adapted here for the tabular domain. Pseudo-labels are derived from an unlabeled pool of \textbf{$u=30,000$} samples and calculated only at the beginning of each session, with the number of high-confidence samples per class set to \textbf{$m=100$}, and are held constant throughout the duration of that session’s training. Comprehensive details regarding hyperparameters and training configurations are provided in Appendix \ref{sec:implementation_details}.

\textbf{Baselines.} We benchmark SPRINT against three categories of competing methods. First, we evaluate classic few-shot learning methods including Prototypical Networks \citep{snell2017prototypical} and MAML \citep{finn2017model}. Second, we compare against established state-of-the-art FSCIL frameworks, specifically FACT \cite{9878986} and iCaRL \cite{Rebuffi_2017_CVPR}. Finally, to isolate the impact of our contributions, we implement two custom adapted baselines: \textit{Semi-Super-ProtoNet}, a semi-supervised variant of ProtoNet that utilizes K-Means clustering for pseudo-label generation, and a tabular adaptation of \textbf{Neuron Expansion} \cite{10720176}. The latter is a transfer-learning approach that dynamically expands the classifier's output layer for each incoming class while fine-tuning the backbone using a balanced replay buffer. Detailed implementation mechanics for Neuron Expansion are provided in Appendix~\ref{sec:neuron_expansion}. We also compared against the self-supervised method STUNT \cite{nam2023stunt}; however, due to poor base session adaptation on these datasets, we detail those results in Appendix ~\ref{sec:additional_analysis}.

\textbf{Evaluation Metrics.} Following the protocol from \cite{9878986}, we evaluate the model using (1) \textbf{Session-wise Accuracy} ($A_i$), which denotes the Top-1 accuracy after session $i$ on all currently observed classes; and (2) \textbf{Performance Dropping Rate (PD)}, which measures the extent of forgetting. PD is calculated as the difference between the base session accuracy ($A_0$) and the accuracy after the final session ($A_S$), i.e., $\text{PD} = A_0 - A_S$.

\begin{table*}[t]
\centering
\caption{Comparison of different models on ACI-IoT-2023 dataset (5-shot and 10-shot). Values indicate mean accuracy (\%). Best results per session are in \textbf{bold}, second-best in \textit{italics}. The `Imp.' column shows the accuracy improvement of SPRINT over each method.}
\begin{center}
\begin{small}
\begin{sc}
\resizebox{\textwidth}{!}{
\begin{tabular}{l|ccccc|c|ccccc|c}
\hline
\multirow{2}{*}{Method} & \multicolumn{6}{c|}{\textbf{5-shot}} & \multicolumn{6}{c}{\textbf{10-shot}} \\ \cline{2-13} 
 & 0 & 1 & 2 & 3 & 4 & Imp. & 0 & 1 & 2 & 3 & 4 & Imp. \\ \hline
CNN ProtoNet & 85.70 & 82.47 & 80.96 & 81.94 & 76.22 & \textbf{+17.42} & 88.25 & 85.33 & 84.49 & 85.14 & 79.91 & \textbf{+15.35} \\
FACT & 77.99 & 76.23 & 74.73 & 75.85 & 69.93 & \textbf{+23.71} & 84.45 & 84.60 & 83.47 & 84.76 & 78.39 & \textbf{+16.88} \\
MAML & 81.18 & 66.98 & 60.06 & 54.87 & 47.27 & \textbf{+46.37} & 83.43 & 68.98 & 61.29 & 55.50 & 47.93 & \textbf{+47.34} \\
Neuron Expansion & \textit{98.11} & 88.58 & 88.38 & 88.05 & 85.79 & \textbf{+7.85} & \textit{98.16} & 92.28 & 90.71 & 91.06 & 89.98 & \textbf{+5.29} \\
ProtoNet & 96.19 & 89.15 & 86.33 & 87.06 & 84.16 & \textbf{+9.47} & 97.23 & 90.86 & 87.57 & 88.08 & 85.25 & \textbf{+10.02} \\
Semi-Super-ProtoNet & 96.17 & 91.01 & 89.18 & 88.84 & 85.13 & \textbf{+8.50} & 97.24 & 92.72 & 90.00 & 89.61 & 85.88 & \textbf{+9.38} \\
iCaRL & \textbf{98.99} & \textbf{97.06} & \textit{95.49} & \textit{94.68} & \textit{89.18} & \textbf{+4.45} & \textbf{98.96} & \textbf{97.43} & \textit{96.59} & \textit{96.28} & \textit{92.04} & \textbf{+3.23} \\
\hline

SPRINT & 96.17 & \textit{96.17} & \textbf{96.29} & \textbf{95.91} & \textbf{93.63} & -- & 97.24 & \textit{96.63} & \textbf{97.02} & \textbf{96.96} & \textbf{95.27} & -- \\
\hline
\end{tabular}
}
\end{sc}
\end{small}
\end{center}
\label{tab:summary_results_sessions_5hot_10shot}
\end{table*}

\subsection{Main Results}\label{sec:main_results}

% We evaluate SPRINT across six diverse datasets, focusing on plasticity (session-wise accuracy) and stability (Performance Dropping rate, PD).

\textbf{State-of-the-Art Performance.} As detailed in Table \ref{tab:summary_results_with_avg}, SPRINT outperforms all baselines in the 5-shot setting, achieving the lowest average PD (\textbf{5.24\%}) and highest final accuracy (\textbf{77.37\%}). On the challenging ACI-IoT-2023 dataset, SPRINT demonstrates exceptional stability with a PD of just \textbf{2.54\%}, outperforming the strongest baseline, iCaRL (PD = 9.81\%). This advantage is visually confirmed in Figure \ref{fig:statistical_sig}, where SPRINT exhibits a tightly clustered accuracy distribution around 93.6\%, contrasting sharply with the high-variance, lower-mean distribution of iCaRL. This improvement is statistically significant ($p < 0.001$, Cohen's $d=2.98$), verifying that our dual-loss objective effectively anchors the embedding space against catastrophic forgetting. Statistical analysis across all datasets further validates that SPRINT's gains over the strongest baseline are significant (family-wise error rate $\alpha=0.05$); for comprehensive hypothesis testing and effect size details, refer to the \textbf{Statistical Significance Analysis} in Appendix \ref{sec:additional_analysis}.

SPRINT proves robust across varying feature properties and architectures. On the high-dimensional \textbf{MNIST} dataset (784 features), using a CNN backbone, SPRINT achieves \textbf{84.85\%} accuracy, surpassing the specialized CNN-ProtoNet (81.10\%). Figure \ref{fig:model_comparison_sessions} illustrates that SPRINT maintains a "flat" performance curve on CIC-IDS2017 and MNIST, avoiding the drops seen in FACT and MAML. Furthermore, Table \ref{tab:summary_results_sessions_5hot_10shot} confirms that this dominance holds regardless of data scarcity; SPRINT consistently outperforms the runner-up (iCaRL) by \textbf{3--4\%} in both 5-shot and 10-shot settings, demonstrating that our prototype refinement does not rely on large support sets. Crucially, while baselines like Neuron Expansion show strong initial performance (Session 0), they fail to match SPRINT's adaptability in later sessions, confirming that structural expansion alone is insufficient without robust prototype maintenance.

% Similarly, on complex tabular datasets like \textbf{Obesity}, SPRINT maintains high stability (PD = 3.61\%) where gradient-based methods like MAML fail (PD = 23.95\%), validating the efficacy of our semi-supervised prototypes in capturing ordinal structures. 

\textbf{Ablation Study.}\label{sec:ablation} Table \ref{tab:ablation_5shot} isolates the contribution of each component on ACI-IoT-2023. Naively applying prototypical loss ($\mathcal{L}_{proto}$) to the Baseline degrades performance (82.64\%) due to overfitting. Adding semi-supervised clustering loss ($\mathcal{L}_{semi-clust.}$) reverses this trend (85.13\%), confirming the value of unlabeled data. The most significant gain is achieved by the full SPRINT framework, marking an \textbf{8.5\% improvement}. This demonstrates that SPRINT's superior performance stems from the \textbf{synergistic interplay} between continuous base class rehearsal (which anchors the embedding space) and \textbf{semi-supervised expansion} (which refines novel class boundaries)

\begin{table}[t]
\caption{Ablation study accuracy (\%) on ACI-IoT-2023 (5-shot).}
\label{tab:ablation_5shot}
\begin{center}
\begin{small}
\begin{sc}
\begin{tabular}{lcccc}
\toprule
Method  & 1 &  2 &  3 &  4 \\
\midrule
Baseline & 89.15 & 86.33 & 87.06 & 84.16 \\
+ $\mathcal{L}_{proto}$  & 88.84 & 85.77 & 86.09 & 82.64 \\
+ $\mathcal{L}_{semi-clust.}$ & 91.01 & 89.18 & 88.84 & 85.13 \\
SPRINT  & \textbf{ 96.17} & \textbf{96.29} & \textbf{95.91} & \textbf{93.63} \\
\bottomrule
\end{tabular}
\end{sc}
\end{small}
\end{center}
\vskip -0.1in
\label{tab:ablations}
\end{table}

\textbf{Architectural Validation and Stability.} We analyze the impact of the distance metric within the prototypical framework. As shown in Figure \ref{fig:distance_comparison}, Euclidean distance consistently outperforms Cosine similarity, achieving an accuracy of 95.27\% on the ACI-IoT-2023 dataset compared to 87.96\%. This aligns with findings by \citet{snell2017prototypical}, who note that Euclidean distance is a Bregman divergence, for which the cluster mean is the optimal representative. In contrast, simple averaging is not the distance-minimizing operation for Cosine similarity, leading to suboptimal cluster representations. Additionally, we evaluated the model in a non-incremental setting (Figure \ref{fig:nway_kshot}), where all novel classes are introduced simultaneously. SPRINT achieves the highest Harmonic Mean among competing models.

\textbf{Hyperparameter Sensitivity.} We analyze the impact of the loss coefficient $\beta$ and pseudo-label pool size $m$. As shown in Figure \ref{fig:betta_ablations}, performance remains stable for $\beta \in [0.4, 0.6]$ but drops sharply to 82.64\% when $\beta=1.0$. This decline confirms that the semi-supervised objective ($\mathcal{L}_{semi}$), which is nullified at $\beta=1.0$, is essential for learning novel classes. Conversely, the model exhibits high robustness to $m$ (Figure \ref{fig:number_of_pseudo_samples}), with accuracy fluctuating marginally (93.62\%--94.28\%) across settings, indicating that filtering for high confidence is more critical than the exact quantity of pseudo-labels.

\textbf{Robustness to Memory Constraints.}\label{sec:robustness} To investigate whether SPRINT's performance is contingent on large memory buffers, we conducted a rigorous sensitivity analysis on the base class memory budget ($M^{(0)}$). We varied the retained samples per base class from $M^{(0)}=4000$ down to severe constraint regimes ($M^{(0)} \in \{ 1000, 500, 100, 50\}$). As illustrated in Figure \ref{fig:memory_budget}, SPRINT maintains a consistent performance advantage over iCaRL and ProtoNets across the entire spectrum, indicating that our dual-loss optimization anchors the latent space efficiently without requiring excessive storage. In terms of data availability, the shot ablation results in Figure \ref{fig:shots} reveals that while the 1-shot setting suffers degradation ($\approx$81\% by Session 4), performance stabilizes rapidly at $K \geq 5$, with near-identical trajectories for 10, 15, and 20-shot settings. This efficiency is underpinned by the high quality of unlabelled data integration; as shown in Figure \ref{fig:pseudo-label-accuracy}, the system generates highly reliable pseudo-labels even in low-data regimes (e.g., Class 7 achieves 98.7\% accuracy), ensuring that the semi-supervised component reinforces correct decision boundaries.

\section{Conclusion}

We presented SPRINT, the first semi-supervised few-shot
class-incremental learning method for tabular domains. By leveraging unlabeled data streams, SPRINT significantly advances the trade-off between stability and plasticity inherent to continuous learning in high-stakes applications. Empirical validation across cybersecurity, healthcare, and ecological benchmarks confirms that SPRINT reduces catastrophic forgetting by over $3\times$ compared to standard baselines. Ultimately, this work \textbf{opens new research directions for continual learning in tabular domain}.

\textbf{Limitations and Future Work.} While our retention strategy is efficient for standard tabular datasets, it may face challenges in strictly regulated environments where data storage is restricted by privacy laws (e.g., HIPAA). Future work will explore privacy-preserving replay mechanisms.

\section*{Acknowledgment}

The research reported herein was supported by Army Research Office (ARO) award W911NF-24-2-0114. The views and opinions presented herein are those of the authors and do not necessarily represent the views of ARO or its components.

\section*{Impact Statement}

\textbf{Societal Benefits.} This paper presents a foundational framework for adaptable learning in tabular streams, with implications in various domains. In cybersecurity, SPRINT enhances the resilience of critical infrastructure by enabling intrusion detection systems to adapt to zero-day attacks rapidly. Beyond defense, this framework holds significant potential for public health, where it could accelerate the identification of emerging pathogens (e.g., novel virus variants) within electronic health records without disrupting diagnostic capabilities for established diseases. Similarly, in environmental monitoring, it enables real-time tracking of shifting ecological patterns from sensor arrays without the computational cost of constant retraining.

\nocite{langley00}

\bibliography{example_paper}
\bibliographystyle{icml2026}

%%%%%%%%%%%%%%%%%%%%%%%%%%%%%%%%%%%%%%%%%%%%%%%%%%%%%%%%%%%%%%%%%%%%%%%%%%%%%%%
%%%%%%%%%%%%%%%%%%%%%%%%%%%%%%%%%%%%%%%%%%%%%%%%%%%%%%%%%%%%%%%%%%%%%%%%%%%%%%%
% APPENDIX
%%%%%%%%%%%%%%%%%%%%%%%%%%%%%%%%%%%%%%%%%%%%%%%%%%%%%%%%%%%%%%%%%%%%%%%%%%%%%%%
%%%%%%%%%%%%%%%%%%%%%%%%%%%%%%%%%%%%%%%%%%%%%%%%%%%%%%%%%%%%%%%%%%%%%%%%%%%%%%%
\newpage
\appendix
\onecolumn

\section{Detailed Computational Complexity Analysis}
\label{sec:complexity}

In this section, we provide a granular breakdown of the computational operations required for SPRINT. We assume a standard MLP encoder architecture $f_\theta: \mathbb{R}^D \to \mathbb{R}^M$ with hidden dimensions $[h_1, h_2, M]$.  Throughout this analysis, $D$ denotes the input feature dimensionality, $M$ denotes the embedding dimensionality, and $|\mathcal{C}^{(\leq t)}|$ denotes the total number of classes seen up to session $t$.

\subsection{Inference Complexity}
At inference time, SPRINT maintains\textbf{ identical computational complexity to standard Prototypical Networks}. For a test sample $x_{\text{test}} \in \mathbb{R}^D$, prediction requires computing the embedding $f_\theta(x_{\text{test}}) \in \mathbb{R}^M$ with cost $O(D \cdot M)$, followed by distance computation $\|f_\theta(x_{\text{test}}) - \mathbf{p}_c\|_2^2$ to all $|\mathcal{C}^{(\leq t)}|$ prototypes with cost $O(|\mathcal{C}^{(\leq t)}| \cdot M)$, and finally classification via argmin with cost $O(|\mathcal{C}^{(\leq t)}|)$. The total inference complexity is therefore $ O(D \cdot M + |\mathcal{C}^{(\leq t)}| \cdot M)$.

This matches standard ProtoNet exactly, as pseudo-labeling only occurs during training. The semi-supervised component introduces \textbf{zero inference overhead}, making SPRINT suitable for production deployment with real-time latency requirements.

\subsection{Training Complexity}
\textbf{Base Session (Session 0):} Both ProtoNet and SPRINT perform identical episodic training with complexity $O(E_{\text{base}} \cdot N_{\text{way}} \cdot (k+q) \cdot D \cdot M)$ for $E_{\text{base}}$ episodes.

\textbf{Incremental Session Phase 1: Pseudo-labeling (once per session).} For each unlabeled sample $x_j \in \mathcal{U}_{\text{batch}}$,
%MK-Done: I removed the  specifci size
%$|\mathcal{U}_{\text{batch}}|$= 30{,}000$, 
we compute embeddings and distances to all $|\mathcal{C}^{(\leq t)}| = |\mathcal{C}^{(0)}| + t$ prototypes, requiring $O(D \cdot M + |\mathcal{C}^{(\leq t)}| \cdot M)$ operations per sample. We then sort distances to select the top-$m$ samples per novel class with complexity $O(|\mathcal{U}_{\text{batch}}| \cdot \log |\mathcal{U}_{\text{batch}}|)$. The total pseudo-labeling cost is:
$
O(|\mathcal{U}_{\text{batch}}| \cdot [(D + |\mathcal{C}^{(\leq t)}|) \cdot M + \log |\mathcal{U}_{\text{batch}}|])
$. Crucially, empirical stability does not rely on $|\mathcal{U}_{\text{batch}}|$ being exhaustively large; in our experiments, a moderate batch size of $|\mathcal{U}_{\text{batch}}| = 30,000$ proved sufficient to capture the manifold structure of novel classes while keeping this step computationally negligible.
%MK-Done: maybe here we say something like U_batch does not need to be too large in practice ??

\textbf{Incremental Session Phase 2: Mixed Episodic Training (for $E_{\text{inc}}$ episodes).} Each episode computes two losses through distinct forward passes. First, we sample $N_{\text{way}}$ base classes with $k$ support and $q$ query samples per class from $\mathcal{M}^{(0)}$, computing $\mathcal{L}_{\text{proto}}^{(t)}$ with cost $O(N_{\text{way}} \cdot (k+q) \cdot D \cdot M)$. Second, we sample $(N_{\text{way}} + |\mathcal{C}^{(t)}|)$ classes (base + novel) from the combined labeled and pseudo-labeled pool, computing $\mathcal{L}_{\text{semi}}^{(t)}$ with cost $O((N_{\text{way}} + |\mathcal{C}^{(t)}|) \cdot (k+q) \cdot D \cdot M)$. Since both sub-episodes execute in every training iteration, the total forward pass cost per episode is $
O((2N_{\text{way}} + |\mathcal{C}^{(t)}|) \cdot (k+q) \cdot D \cdot M)$. The combined loss $\mathcal{L}^{(t)} = \beta \cdot \mathcal{L}_{\text{proto}}^{(t)} + (1-\beta) \cdot \mathcal{L}_{\text{semi}}^{(t)}$ requires one backward pass with complexity matching the forward pass. For $E_{\text{inc}}$ episodes, Phase 2 complexity is $O(E_{\text{inc}} \cdot (2N_{\text{way}} + |\mathcal{C}^{(t)}|) \cdot (k+q) \cdot D \cdot M)$.

\textbf{Total Incremental Session Complexity:}
\begin{equation*}
O(|\mathcal{U}_{\text{batch}}| \cdot |\mathcal{C}^{(\leq t)}| \cdot M + E_{\text{inc}} \cdot (2N_{\text{way}} + |\mathcal{C}^{(t)}|) \cdot (k+q) \cdot D \cdot M)
\end{equation*}

\subsection{Forward and Backward Pass Operations}

\textbf{Encoder Forward Pass Cost ($C_{fwd}$).} For a single input sample $x \in \mathbb{R}^D$, the forward pass through a 3-layer MLP involves matrix-vector multiplications and additions:
\begin{align}
\text{Layer 1:} \quad &D \cdot h_1 \text{ mults} + h_1 \text{ adds} \nonumber \\
\text{Layer 2:} \quad &h_1 \cdot h_2 \text{ mults} + h_2 \text{ adds} \nonumber \\
\text{Layer 3:} \quad &h_2 \cdot M \text{ mults} + M \text{ adds} \nonumber
\end{align}
Assuming $h_1 \approx h_2 \approx M$ for simplicity, the dominant complexity is:
\begin{equation}
C_{fwd} \approx O(D \cdot M + M^2) \approx O(D \cdot M) \quad (\text{since } D \ge M)
\end{equation}

\textbf{Encoder Backward Pass Cost ($C_{bwd}$).} Using standard automatic differentiation (backpropagation), the computational cost is proportional to the forward pass:
\begin{equation}
C_{bwd} \approx 2 \cdot C_{fwd} = O(D \cdot M)
\end{equation}

\textbf{Prototype Computation ($C_{proto}$).} Computing a prototype for class $c$ by averaging $k$ support embeddings requires $k \cdot M$ additions and $M$ divisions:
\begin{equation}
C_{proto}(k) = O(k \cdot M)
\end{equation}

\textbf{Distance \& Loss Computation ($C_{dist}$).} For a query set of size $N_Q$ and a support set of $N_C$ classes:
\begin{itemize}
    \item Pairwise Euclidean distances: $O(N_Q \cdot N_C \cdot M)$
    \item Softmax \& Log-likelihood: $O(N_Q \cdot N_C)$
\end{itemize}

\subsection{Phase 1: Pseudo-labeling Detailed Breakdown}

This phase occurs once at the beginning of each incremental session $t$.

\textbf{Step 1: Batch Forward Pass.} We process the entire unlabeled batch $\mathcal{U}_{\text{batch}}$ to generate embeddings.
\begin{equation}
T_{step1} = |\mathcal{U}_{\text{batch}}| \cdot C_{fwd} = O(|\mathcal{U}_{\text{batch}}| \cdot D \cdot M)
\end{equation}

\textbf{Step 2: Distance Matrix Calculation.} We compute distances between every unlabeled sample and all existing prototypes $\mathcal{C}^{(\leq t)}$.
\begin{equation}
T_{step2} = |\mathcal{U}_{\text{batch}}| \cdot |\mathcal{C}^{(\leq t)}| \cdot M
\end{equation}

\textbf{Step 3: Top-$m$ Selection.} For each novel class $c \in \mathcal{C}^{(t)}$, we select the top-$m$ high-confidence samples. This involves a partial sort (e.g., via a heap or introselect) of the distance vector of size $|\mathcal{U}_{\text{batch}}|$.
\begin{equation}
T_{step3} = |\mathcal{C}^{(t)}| \cdot O(|\mathcal{U}_{\text{batch}}| \cdot \log m)
\end{equation}

\textbf{Total Phase 1 Complexity:}
\begin{equation}
T_{phase1} \approx O\Big(|\mathcal{U}_{\text{batch}}| \cdot \big(D \cdot M + |\mathcal{C}^{(\leq t)}| \cdot M + |\mathcal{C}^{(t)}|\log m\big)\Big)
\end{equation}
Given that $|\mathcal{C}^{(t)}| \ll D$ and $\log m$ is negligible, this is dominated by the embedding generation: $O(|\mathcal{U}_{\text{batch}}| \cdot D \cdot M)$.

\subsection{Phase 2: Mixed Episodic Training Detailed Breakdown}

For each of the $E_{\text{inc}}$ training episodes, we perform two sub-episodes involving forward and backward passes.

\textbf{Sub-episode 1: Base Rehearsal ($\mathcal{L}_{proto}^{(t)}$).} This phase utilizes data from $N_{\text{way}}$ classes, each comprising $k$ support and $q$ query samples, for a total of $S_1 = N_{\text{way}}(k+q)$ samples. The forward pass embeds these $S_1$ samples with a computational cost of $S_1 \cdot C_{fwd}$. Subsequently, prototype construction involves averaging $N_{\text{way}}$ clusters at a cost of $N_{\text{way}} \cdot k \cdot M$. Finally, distance calculation is performed between the $N_{\text{way}} \cdot q$ query samples and the $N_{\text{way}}$ prototypes, incurring a cost of $N_{\text{way}}^2 \cdot q \cdot M$.

\textbf{Sub-episode 2: Semi-Supervised Learning ($\mathcal{L}_{semi}^{(t)}$).} This phase expands the scope to $(N_{\text{way}} + |\mathcal{C}^{(t)}|)$ classes, yielding a total sample size of $S_2 = (N_{\text{way}} + |\mathcal{C}^{(t)}|)(k+q)$. The forward pass embeds these $S_2$ samples with a cost of $S_2 \cdot C_{fwd}$. Prototype construction then averages the $(N_{\text{way}} + |\mathcal{C}^{(t)}|)$ clusters at a cost of $(N_{\text{way}} + |\mathcal{C}^{(t)}|) \cdot k \cdot M$. The process concludes with distance calculations between queries and prototypes, with a complexity of $(N_{\text{way}} + |\mathcal{C}^{(t)}|)^2 \cdot q \cdot M$.

\subsection{Phase 2: Mixed Episodic Training Detailed Breakdown}

For each of the $E_{\text{inc}}$ training episodes, we perform two sub-episodes involving forward and backward passes.

\textbf{Backward Pass (Joint Optimization).}
We perform backpropagation on the weighted loss $\mathcal{L} = \beta \mathcal{L}_{proto} + (1-\beta) \mathcal{L}_{semi}$. This requires gradients for all samples processed in both sub-episodes.
\begin{equation}
T_{backward} = (S_1 + S_2) \cdot C_{bwd} = O\Big((2N_{\text{way}} + |\mathcal{C}^{(t)}|)(k+q) \cdot D \cdot M\Big)
\end{equation}

\textbf{Total Phase 2 Complexity (Per Episode):}
Summing forward, backward, and metric computations, the per-episode cost is dominated by the embedding of samples:
\begin{equation}
T_{episode} \approx O\Big((2N_{\text{way}} + |\mathcal{C}^{(t)}|) \cdot (k+q) \cdot D \cdot M\Big)
\end{equation}

\subsection{Empirical Runtime Analysis}
\label{sec:empirical_runtime}

To validate our complexity analysis, we profiled the wall-clock training time of the incremental learning phase on the ACI-IoT-2023 dataset. We compared SPRINT against iCaRL \cite{Rebuffi_2017_CVPR} under identical experimental conditions: a fixed memory budget of $|\mathcal{M}^{(0)}| = 2,000$ samples per class and the same hardware configuration (NVIDIA L40S).

\textbf{Infrastructure Setup.} All experiments were conducted on a high-performance computing node equipped with 8 $\times$ NVIDIA L40S GPUs (48 GB GDDR6 VRAM per accelerator)  running on Driver Version 580.95.05 (CUDA 13.0). To ensure a fair comparison and isolate algorithmic efficiency from multi-GPU parallelism, the runtime profiling utilized a \textbf{single GPU instance}. The software environment was configured with Python 3.12.6, PyTorch 2.4.1 (CUDA 12.1 runtime), Numpy 2.1.1, and Scikit-learn 1.5.2.

\begin{figure}[h]
    \centering
    \includegraphics[width=0.6\linewidth]{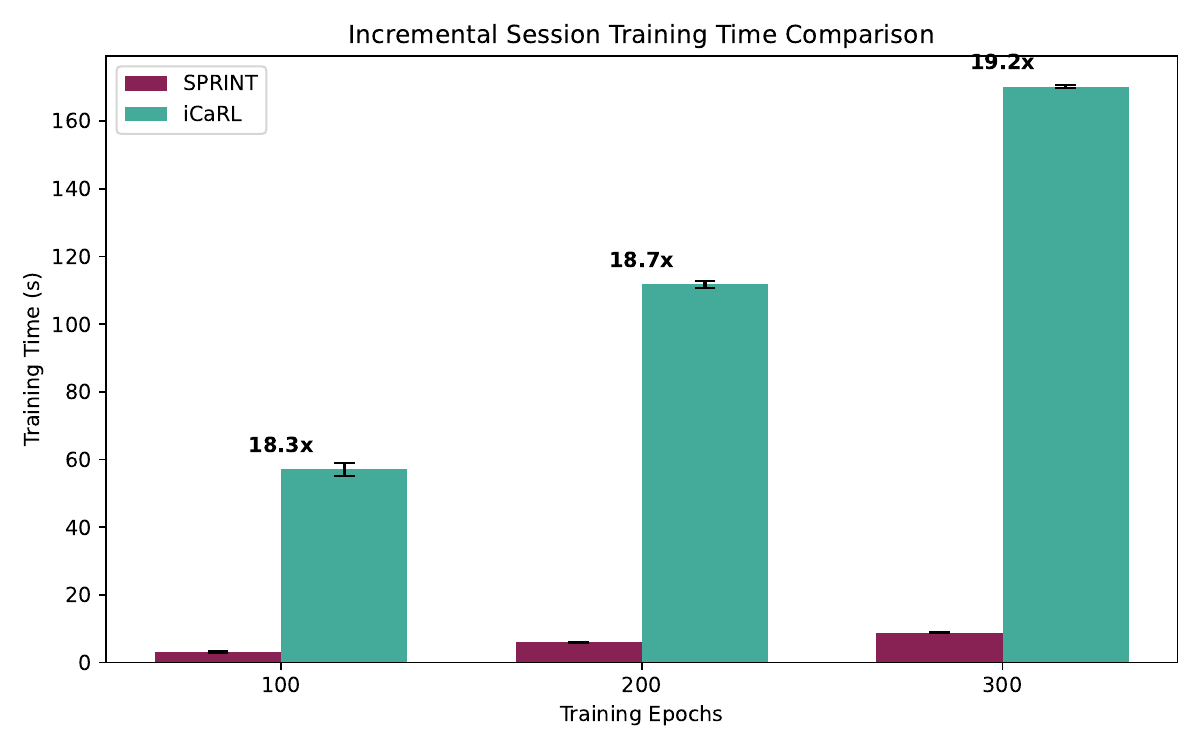} 
    \caption{\textbf{Incremental Training Latency Comparison.} Training runtime (in seconds) averaged across all incremental sessions and 10 independent runs. Numbers above bars indicate the speedup factor ($T_{\text{iCaRL}} / T_{\text{SPRINT}}$). SPRINT consistently achieves an $\approx 18\times$ reduction in training latency due to its sparse episodic sampling strategy, which decouples computational cost from the total memory buffer size.}
    \label{fig:runtime_comparison}
\end{figure}

\begin{table}[h]
\centering
\caption{Training time profiling (seconds). Results averaged over 10 independent runs ($\pm$ std dev).}
\label{tab:runtime_data}
\begin{small}
\begin{sc}
\begin{tabular}{lcccc}
\toprule
\textbf{Method} & \textbf{100 Epochs} & \textbf{200 Epochs} & \textbf{300 Epochs} & \textbf{Speedup} \\
\midrule
iCaRL & $57.09 \pm 1.87$ & $111.78 \pm 0.98$ & $170.21 \pm 0.48$ & $1.0\times$ \\
SPRINT & $\mathbf{3.12} \pm 0.15$ & $\mathbf{5.97} \pm 0.14$ & $\mathbf{8.87} \pm 0.17$ & $\mathbf{18.3\times}$ \\
\bottomrule
\end{tabular}
\end{sc}
\end{small}
\end{table}

As illustrated in Figure \ref{fig:runtime_comparison}, SPRINT exhibits an order-of-magnitude efficiency gain. For a standard training duration of 100 epochs, iCaRL requires 57.1s, whereas SPRINT completes the optimization in just 3.1s ($18.3\times$ speedup). This discrepancy arises because iCaRL's dense replay mechanism forces a full pass over the expanded memory buffer ($|\mathcal{M}^{(0)}| \cdot N_{way}$) in every epoch. In contrast, SPRINT's episodic strategy samples small batches ($N_{way} \cdot (k+q)$) from memory, keeping per-iteration costs constant and low regardless of the total buffer size. Table \ref{tab:runtime_data} provides the detailed measurements.

\subsection{Memory Complexity}

\textbf{Storage ($\mathcal{M}^{(0)}$).} Unlike image domains where retaining raw pixels is prohibitive, storing tabular base data is highly efficient. For $M^{(0)}$ samples per class and $|\mathcal{C}^{(0)}|$ base classes:
\begin{equation}
\text{Storage} = |\mathcal{C}^{(0)}| \cdot M^{(0)} \cdot D \cdot \text{sizeof(float)}
\end{equation}
For ACI-IoT-2023 ($D=79$, $|\mathcal{C}^{(0)}|=6$, $M^{(0)}=2000$), this requires $\approx 3.8$ MB, which is negligible for modern hardware.

\section{Detailed Class Partitions and Split Configurations}
\label{sec:data_splits} 

In this section, we provide the specific experimental configurations, class partitions, and data splits used to benchmark SPRINT against state-of-the-art methods. To ensure a rigorous Few-Shot Class-Incremental Learning (FSCIL) evaluation, we categorize the classes of each dataset into two disjoint sets: 
\begin{itemize}
    \item \textbf{Base Classes ($C_{base}$):} Classes available during the initial offline training phase (Session 0).
    \item \textbf{Novel Classes ($C_{novel}$):} Classes introduced sequentially in subsequent incremental sessions (1 class per session).
\end{itemize}

The partitions are stratified to ensure semantic separation between base knowledge and novel concepts. We utilize a fixed number of samples per class ($N_{source}$) for the base training to simulate varying degrees of data availability.

\subsection*{A.1 Cybersecurity Datasets}
\begin{itemize}
    \item \textbf{ACI-IoT-2023} \cite{qacj-3x32-23}: We utilize the 2023 IoT dataset to evaluate performance on modern attack vectors. The base session is constructed using 6 classes, comprising \textit{Benign} traffic and surveillance-related attacks (e.g., \textit{OS Scan}, \textit{Port Scan}). The incremental phase introduces 4 distinct flooding attacks (\textit{ICMP Flood}, \textit{Slowloris}, etc.) to test the model's adaptability to high-volume traffic anomalies.
    
    \item \textbf{CIC-IDS2017} \cite{sharafaldin2018toward}: A standard network intrusion detection benchmark. We select 5 classes for the base session, covering normal traffic and brute-force activities (\textit{FTP/SSH Patator}). The incremental sessions introduce 4 variants of Denial of Service (DoS) attacks, requiring the model to distinguish between subtle variations of volumetric attacks.
    
    \item \textbf{CIC-IoT2023} \cite{neto2023ciciot2023}: To test scalability, we implement a large-scale split on this dataset. The base session includes 14 diverse classes, ranging from \textit{Mirai} botnet activities to reconnaissance tactics. The incremental phase is extensive, introducing 11 novel classes focused primarily on various \textit{DDoS} implementations.
\end{itemize}

\subsection*{A.2 Tabular \& Medical Datasets}
\begin{itemize}
    \item \textbf{Obesity} \cite{Palechor2019DatasetFE}: This dataset categorizes health risk based on physical attributes and habits. We create an ordinal split where "lower risk" categories (\textit{Insufficient Weight}, \textit{Normal}, \textit{Overweight}) form the base knowledge ($|C_{base}|=4$). The higher-risk categories (\textit{Obesity Type I, II, III}) are introduced as novel classes ($|C_{novel}|=3$), testing the model's ability to learn severity progression.
    
    \item \textbf{CovType} \cite{covertype_31}: We utilize the standard forest cover type classification problem. The first 4 cover types (\textit{Spruce/Fir, Lodgepole Pine, Ponderosa Pine, Cottonwood/Willow}) are used for base training. The remaining 3 types (\textit{Aspen, Douglas-fir, Krummholz}) are introduced incrementally.

    \item \textbf{MNIST} \cite{lecun2010mnist}: To benchmark our architecture on high-dimensional pattern recognition tasks, we utilize the MNIST dataset in a flattened tabular format. Each $28 \times 28$ pixel grayscale image is treated as a 1D feature vector, resulting in a high-dimensionality input space of 784 features ($\mathbb{R}^{784}$). This setup specifically tests the model's ability to extract meaningful prototypes from sparse, high-dimensional tabular data. We utilize the first 6 digits (0--5) for base training and introduce the remaining digits (6--9) incrementally.
\end{itemize}

\section{Appendix: Implementation Details}
\label{sec:implementation_details}

To ensure reproducibility and statistical rigor, we detail the hyperparameters and evaluation protocols used in our experiments in Table \ref{tab:implementation_details}.

\textbf{Experimental Protocol.}
Evaluation results are reported as the mean accuracy over \textbf{30 independent runs}. Within each run, we sample \textbf{500 randomly generated test episodes} to minimize variance and ensure that reported confidence intervals reflect true model stability. Following standard few-shot protocols~\cite{snell2017prototypical}, the size of the support set is fixed to match the shot number $K$ (e.g., in a 5-shot experiment, the support set contains 5 examples per class).

\textbf{Hyperparameter Configuration.}
Our dual-loss objective is balanced using a coefficient $\beta = 0.5$, giving equal weight to the prototypical and semi-supervised components. The model is trained for \textbf{300 episodes} per session with a learning rate of \textbf{0.001}. For the semi-supervised component, we utilize an unlabeled dataset size of $u = 30,000$ to ensure sufficient diversity in the transductive inference stage. The number of query points is set to 15 per class. Regarding the pseudo-labeling mechanism, we set the maximum number of high-confidence samples added to the support set, $m$, to \textbf{100}. This threshold was empirically selected to strike a balance between computational efficiency and the reliability of the augmented support set; increasing $m$ beyond this point yields diminishing returns while increasing the risk of introducing noisy labels into the prototypical cluster computation.

\begin{table}[h]
\centering
\caption{Summary of implementation hyperparameters and evaluation protocols.}
\label{tab:implementation_details}
\begin{center}
%\begin{small}
%\begin{sc}
\begin{tabular}{l c l}
\toprule
\textbf{Parameter} & \textbf{Value} & \textbf{Description} \\
\midrule
\multicolumn{3}{l}{\textit{Evaluation Protocol}} \\
Independent Runs & 30 & Number of distinct random seeds used for reporting means \\
Test Episodes & 500 & Number of episodes evaluated per run \\
Support Set Size & $K$ & Matches the $K$-shot setting~\cite{snell2017prototypical} \\
Query Points ($N_Q$) & 15 & Number of query samples per class per episode \\
\midrule
\multicolumn{3}{l}{\textit{Training \& Optimization}} \\
Training Episodes & 300 & Episodes per incremental session \\
Learning Rate & 0.001 & Initial learning rate for the optimizer \\
$\beta$ (Balancing Term) & 0.5 & Weight for $\mathcal{L}_{semi}$ vs $\mathcal{L}_{proto}$ \\
\midrule
\multicolumn{3}{l}{\textit{Semi-Supervised Component}} \\
Unlabeled Set Size ($u$) & 30,000 & Size of the available unlabeled pool \\
Pseudo-label Limit ($m$) & 100 & Max high-confidence samples added per class \\
\bottomrule
\end{tabular}%
%\end{sc}
%\end{small}
\end{center}
\end{table}

\section{Implementation Details of Neuron Expansion}
\label{sec:neuron_expansion}

\begin{figure}[t]
    \centering
    \includegraphics[width=0.4\linewidth]{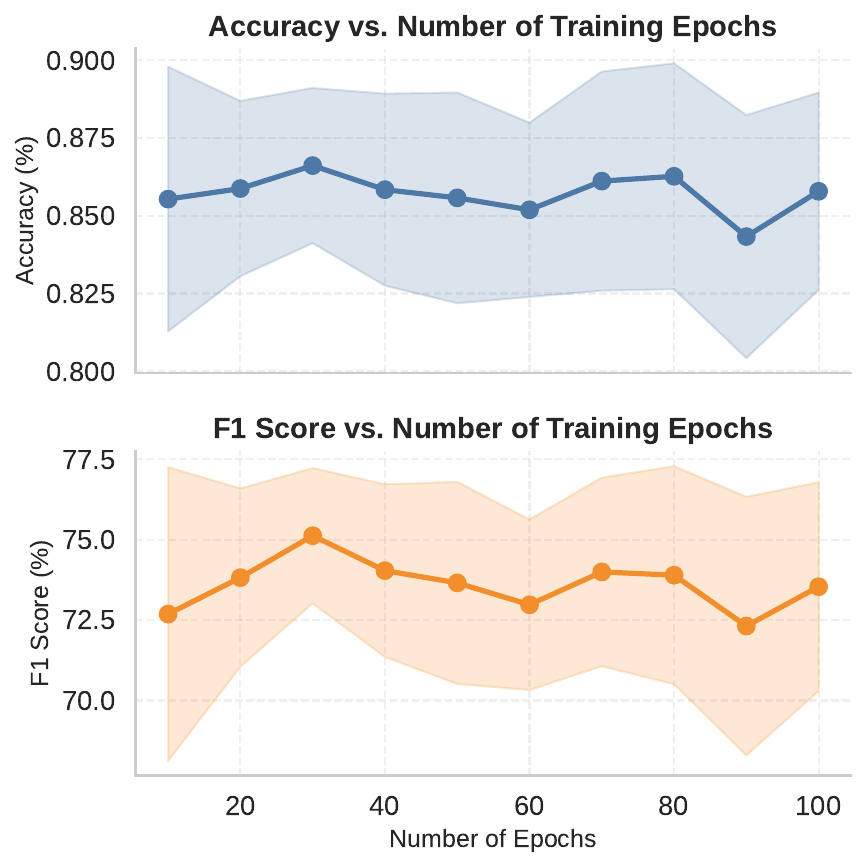}
    \caption{\textbf{Impact of Training Duration on Neuron Expansion.} Performance sensitivity (Top-1 Accuracy and F1 Score) of the Neuron Expansion model across varying epoch counts on the ACI-IoT-2023 dataset. The model achieves peak stability around 30 epochs, after which performance plateaus or slightly degrades, indicating limited benefit from prolonged training.}
    \label{fig:neuron_expansion_epochs}
\end{figure}

Following the transfer-learning paradigm described in \cite{10720176}, we implement \textbf{Neuron Expansion} as a dynamic architectural baseline. Unlike static classifiers, this method physically modifies the network structure during the incremental adaptation phase ($T_1$) to accommodate novel concepts.

\textbf{Architecture.} 
Consistent with our proposed method, the backbone $f_\theta(\cdot)$ is a 3-layer MLP with 1,024 hidden units and ReLU activations. The final classification layer $\phi(\cdot)$ is dynamic, growing with the number of observed classes.

\textbf{Training Protocol.}
The learning process is divided into two phases:
\begin{itemize}
    \item \textbf{Pre-training ($T_0$):} The model is initially trained on the disjoint base set $C_{base}$ using standard Cross-Entropy loss. Let $\mathbf{W}_{old}$ and $\mathbf{b}_{old}$ denote the weights and biases of the classifier $\phi_{old}$ after this phase.
    
    \item \textbf{Incremental Adaptation ($T_1$):} Upon the arrival of a novel class $c_{new}$, we apply the following expansion strategy:
    \begin{enumerate}
        \item \textbf{Structural Expansion:} We expand the output layer dimension from $|C_{old}|$ to $|C_{old}| + 1$. A new neuron is appended to the classifier $\phi$, initialized with random weights, while the existing weights corresponding to previously learned classes are explicitly copied from the previous session to preserve historical knowledge:
        \begin{equation}
            \mathbf{W}_{new}[1:|C_{old}|] \leftarrow \mathbf{W}_{old}, \quad \mathbf{b}_{new}[1:|C_{old}|] \leftarrow \mathbf{b}_{old}
        \end{equation}
        \item \textbf{Balanced Fine-Tuning:} To mitigate class imbalance between the abundant base classes and the scarce novel class (few-shot), we construct a balanced mini-batch. We sample $K$ support examples from the new class and replay $K$ examples randomly sampled from the stored source data.
        \item \textbf{Optimization:} The full network $\langle \theta, \phi \rangle$ is fine-tuned using the Adam optimizer ($\eta=0.001$) to integrate the new decision boundary without catastrophically disrupting the feature embedding space.
    \end{enumerate}
\end{itemize}

To determine the optimal training duration for the parameter isolation baseline, we analyzed the sensitivity of the Neuron Expansion model to the number of training epochs. As illustrated in Figure \ref{fig:neuron_expansion_epochs}, we tracked both Top-1 Accuracy and F1 Score on the ACI-IoT-2023 dataset (Session 4) across a range of 10 to 100 epochs. The results indicate that performance saturates relatively quickly; the model reaches near-peak F1 scores ($\approx$75\%) by 30 epochs. Extending training beyond this point (e.g., to 80 or 100 epochs) yields negligible gains and, in some cases, minor degradation due to overfitting on the limited support set. Consequently, we selected an early stopping point for Neuron Expansion to balance computational efficiency with generalization performance

% \begin{table}[t]
% \caption{Ablation study accuracy (\%) on ACI-IoT-2023 (5-shot) with standard deviations.}
% \label{tab:ablation_5shot_std}
% \begin{center}
% \begin{small}
% \begin{sc}
% \begin{tabular}{lccccc}
% \toprule
% Method & Session 0 & Session 1 & Session 2 & Session 3 & Session 4 \\
% \midrule
% Baseline & 96.19 $\pm$ 0.85 & 89.15 $\pm$ 1.01 & 86.33 $\pm$ 1.26 & 87.06 $\pm$ 1.28 & 84.16 $\pm$ 1.21 \\
% + $\mathcal{L}_{proto}$ & 96.17 $\pm$ 0.87 & 88.84 $\pm$ 1.47 & 85.77 $\pm$ 1.55 & 86.09 $\pm$ 1.84 & 82.64 $\pm$ 1.91 \\
% + $\mathcal{L}_{semi}$ & 96.17 $\pm$ 0.87 & 91.01 $\pm$ 1.20 & 89.18 $\pm$ 1.53 & 88.84 $\pm$ 1.48 & 85.13 $\pm$ 2.15 \\
% SPRINT (Ours) & 96.17 $\pm$ 0.87 & 96.17 $\pm$ 0.87 & 96.29 $\pm$ 0.82 & 95.91 $\pm$ 0.91 & 93.63 $\pm$ 0.88 \\
% \bottomrule
% \end{tabular}
% \end{sc}
% \end{small}
% \end{center}
% \vskip -0.1in
% \end{table}

\begin{table*}[t]
\centering
\caption{Summary of results on 6 datasets (10-shot setting). PD: Performance Dropping rate ($A_0 - A_{last}$), Acc: Last session accuracy ($A_{last}$). Avg-PD and Avg-Acc represent the average values across all datasets. Best results are in \textbf{bold}, second best in \textit{italics}.}
\resizebox{\textwidth}{!}{
\begin{tabular}{l|cc|cc|cc|cc|cc|cc|cc}
\hline
\multirow{2}{*}{Method} & \multicolumn{2}{c|}{ACI-IoT-2023} & \multicolumn{2}{c|}{Obesity} & \multicolumn{2}{c|}{CICIDS2017} & \multicolumn{2}{c|}{CIC-IoT-2023} & \multicolumn{2}{c|}{CovType} & \multicolumn{2}{c|}{MNIST} & \multicolumn{2}{c}{\textbf{Average}} \\
  & PD & Acc & PD & Acc & PD & Acc & PD & Acc & PD & Acc & PD & Acc & \textbf{Avg-PD }& \textbf{Avg-Acc} \\
\hline
CNN ProtoNet & 8.33 & 79.91 & \textit{5.53} & 60.32 & 21.02 & 73.34 & 0.68 & 56.17 & \textbf{9.39} & 49.21 & \textit{7.11} & \textit{86.09} & \textit{8.68} & 67.51 \\
FACT & \textit{6.06} & 78.39 & 11.15 & 15.19 & 13.29 & 18.71 & \textit{-3.04} & 53.83 & 12.70 & 16.40 & 22.63 & 40.59 & 10.47 & 37.19 \\
MAML & 35.50 & 47.93 & 26.13 & 38.62 & 43.53 & 43.31 & 19.44 & 34.44 & 32.86 & 39.19 & 44.17 & 49.89 & 33.61 & 42.23 \\
Neuron Expansion & 8.18 & 89.98 & 14.40 & 67.74 & 17.54 & 77.22 & 28.77 & 62.15 & 28.60 & 51.94 & 19.40 & 78.68 & 19.48 & 71.29 \\
ProtoNet & 11.98 & 85.25 & 10.35 & 72.71 & 26.17 & 70.55 & 2.45 & 61.40 & 10.94 & 57.24 & 22.27 & 73.23 & 14.03 & 70.06 \\

Semi-Super-ProtoNet & 11.36 & 85.88 & 13.14 & 69.95 & 27.15 & 69.59 & 0.35 & 63.49 & 10.53 & \textit{57.79} & 22.79 & 72.70 & 14.22 & 69.90 \\
iCaLR & 6.92 & \textit{92.04} & 5.88 & \textit{79.27} & \textit{13.15} & \textit{84.91} & 5.34 & \textbf{69.51} & 29.48 & 57.68 & 21.15 & 76.93 & 13.66 & \textit{76.72} \\
\hline
SPRINT & \textbf{1.97} & \textbf{95.27} & \textbf{-0.84} & \textbf{83.92} & \textbf{8.77} & \textbf{87.96} & \textbf{-3.97} & \textit{67.82} & \textit{10.26} & \textbf{61.37} & \textbf{3.91} & \textbf{89.32} & \textbf{3.35} & \textbf{80.94} \\
\hline
\end{tabular}
}
\label{tab:summary_results_with_avg_10_shot}
\end{table*}

\section{Additional Analysis}
\label{sec:additional_analysis}

\subsection{Statistical Significance Analysis}
\label{sec:significance}

\begin{figure}[t]
    \centering
    \includegraphics[width=0.8\linewidth]{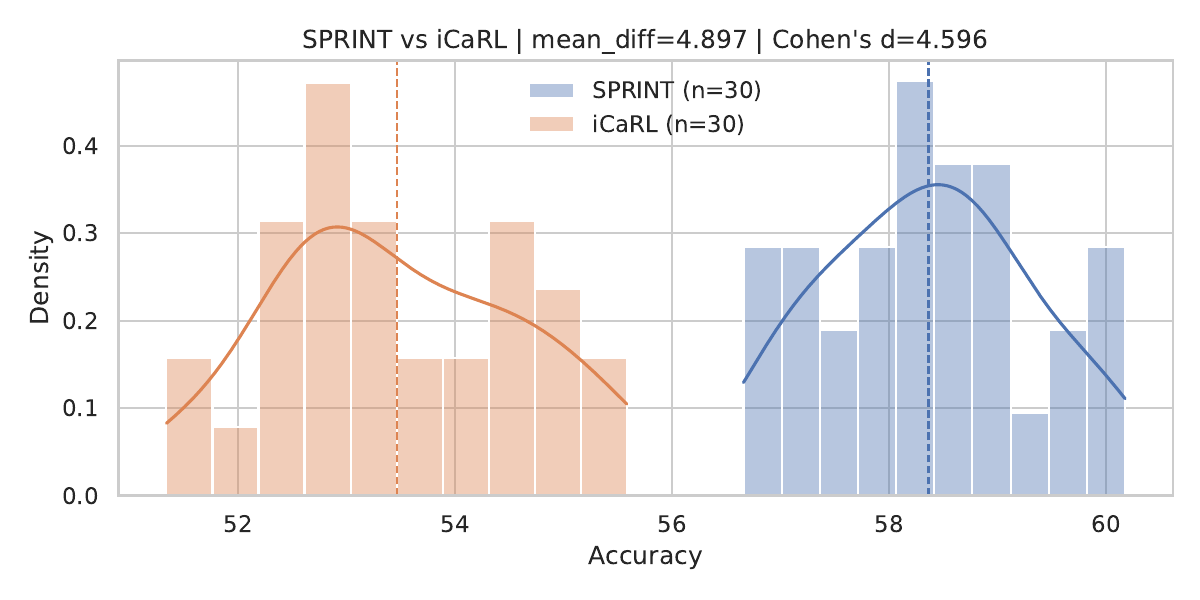}
    \caption{\textbf{Distribution of Top-1 Accuracy (Final Session, $t=4$).} Kernel Density Estimate (KDE) and histogram comparison of SPRINT vs. iCaRL over 30 independent runs on the \textbf{CovType} dataset. SPRINT shows a substantial performance shift ($\Delta \approx 4.9\%$) with a large effect size ($d=4.60$), exhibiting a distribution that is completely separated from the baseline.}
    \label{fig:statistical_sig_cov_type}
\end{figure}

To rigorously validate the performance improvements of SPRINT, we conducted statistical significance tests based on \textbf{30 independent evaluation runs}. For each dataset, we compared the final session accuracy of SPRINT against competing models using a one-sided Welch’s $t$-test (unequal variances), testing the hypothesis that SPRINT's mean accuracy is strictly greater than that of the baseline ($H_1: \mu_{SPRINT} > \mu_{baseline}$). To control the family-wise error rate (FWER) due to multiple comparisons, we applied the \textbf{Bonferroni correction} \citep{dunn1961multiple}. We also report \textbf{Cohen’s $d$} \citep{cohen2013statistical} to quantify the effect size, classifying the magnitude of improvement.

\textbf{Analysis on ACI-IoT-2023.}
Table \ref{tab:significance_aci} details the pairwise comparisons for the ACI-IoT-2023 dataset. SPRINT demonstrates statistically significant superiority over all competitors at the 95\% confidence level ($\alpha=0.05$). The improvement is most pronounced against meta-learning approaches like MAML ($\Delta \approx 46\%$, $d=23.14$) and standard ProtoNets ($\Delta \approx 9.5\%$, $d=8.94$). Crucially, SPRINT also significantly outperforms the strongest incremental baseline, iCaRL, with a mean improvement of \textbf{4.46\%} and a large effect size ($d=2.98$, $p < 0.001$).

\textbf{Visualizing Cross-Domain Stability.}
While the main text illustrates the stability of SPRINT on network traffic data (Figure \ref{fig:statistical_sig}), we provide further evidence of cross-domain robustness in Figure \ref{fig:statistical_sig_cov_type} using the \textbf{CovType} dataset. Here, the performance gap is even more distinct: SPRINT achieves a mean improvement of \textbf{4.897\%} over iCaRL with an exceptionally large effect size (\textbf{$d=4.60$}). The KDE plots reveal two entirely disjoint distributions, indicating that SPRINT outperforms the baseline in every single random trial ($n=30$), thereby confirming that our dual-loss optimization provides consistent gains regardless of seed initialization or data shuffling.

\begin{table}[h]
\centering
\caption{Bonferroni-corrected one-sided p-values and effect sizes (Cohen’s $d$) for SPRINT vs. competing models on ACI-IoT-2023 (5-shot, Final Session). Results are based on 30 independent runs. \textbf{Bold} indicates statistical significance ($p < 0.05$).}
\label{tab:significance_aci}
\begin{center}
\begin{small}
\begin{sc}
\begin{tabular}{llrrcl}
\toprule
\textbf{Model} & \textbf{Compared To} & \textbf{Mean $\Delta$ (\%)} & \textbf{Cohen's $d$} & \textbf{$p$ (Bonf.)} & \textbf{Sig.} \\
\midrule
SPRINT & MAML & \textbf{46.366} & \textbf{23.14} & \textbf{\(<\)0.001} & Yes \\
SPRINT & FACT & \textbf{23.707} & \textbf{1.88} & \textbf{\(<\)0.001} & Yes \\
SPRINT & CNN ProtoNet & \textbf{17.417} & \textbf{12.40} & \textbf{\(<\)0.001} & Yes \\
SPRINT & ProtoNet & \textbf{9.471} & \textbf{8.94} & \textbf{\(<\)0.001} & Yes \\
SPRINT & Semi-Super-Proto & \textbf{8.502} & \textbf{5.18} & \textbf{\(<\)0.001} & Yes \\
SPRINT & Neur. Exp. & \textbf{7.846} & \textbf{3.32} & \textbf{\(<\)0.001} & Yes \\
SPRINT & iCaRL & \textbf{4.455} & \textbf{2.98} & \textbf{\(<\)0.001} & Yes \\
\bottomrule
\end{tabular}%
\end{sc}
\end{small}
\end{center}
\end{table}

\begin{table}[h]
\centering
\caption{Summary of statistical significance tests comparing \textbf{SPRINT vs. iCaRL} (the strongest baseline) across all datasets. Positive $\Delta$ indicates SPRINT outperforms iCaRL.}
\label{tab:significance_summary}
\begin{center}
\begin{small}
\begin{sc}
\begin{tabular}{l rr c c}
\toprule
\textbf{Dataset} & \textbf{Mean $\Delta$ (\%)} & \textbf{Cohen's $d$} & \textbf{$p$ (Bonf.)} & \textbf{Sig.} \\
\midrule
\textbf{ACI-IoT-2023} & \textbf{4.455} & \textbf{2.98} & \textbf{\(<\)0.001}& Yes \\
\textbf{CIC-IDS2017} & \textbf{2.571} & \textbf{1.11} & \textbf{\(<\)0.001}& Yes \\
\textbf{CIC-IoT-2023} & -5.421 & -4.02 & 1.00 & No \\
\textbf{CovType} & \textbf{4.897} & \textbf{4.60} & \textbf{\(<\)0.001} & Yes \\
\textbf{Obesity} & \textbf{3.959} & \textbf{1.97} & \textbf{\(<\)0.001} & Yes \\
\textbf{MNIST} & \textbf{13.852} & \textbf{6.96} & \textbf{\(<\)0.001} & Yes \\
\bottomrule
\end{tabular}%
\end{sc}
\end{small}
\end{center}
\end{table}

\begin{table}
\caption{Bonferroni-corrected one-sided p-values and effect sizes (Cohen’s d) vs SPRINT on 5-shot last-session accuracies (CICIDS2017). All tests conducted at a 95\% confidence level.}
\label{tab:df_significance_CICIDS2017}
\begin{center}
\begin{small}
\begin{sc}
\begin{tabular}{llrrcl}
\toprule
Model & Compared To & Mean $\Delta$ (\%) & Cohen's d & p (Bonf.) & Signif. (Bonf.) \\
\midrule
SPRINT & MAML & \textbf{40.129} & \textbf{16.57} & \textbf{\(<\)0.001} & Yes \\
SPRINT & FACT & \textbf{64.381} & \textbf{17.88} & \textbf{\(<\)0.001} & Yes \\
SPRINT & Semi-Super-ProtoNet & \textbf{15.332} & \textbf{6.70} & \textbf{\(<\)0.001} & Yes \\
SPRINT & ProtoNet & \textbf{13.593} & \textbf{5.59} & \textbf{\(<\)0.001} & Yes \\
SPRINT & CNN ProtoNet & \textbf{12.892} & \textbf{5.72} & \textbf{\(<\)0.001} & Yes \\
SPRINT & Neuron Expansion & \textbf{8.790} & \textbf{1.87} & \textbf{\(<\)0.001} & Yes \\
SPRINT & iCaRL & \textbf{2.571} & \textbf{1.11} & \textbf{\(<\)0.001} & Yes \\
\bottomrule
\end{tabular}
\end{sc}
\end{small}
\end{center}
\end{table}

\begin{figure}[t]
    \centering
    \includegraphics[width=\textwidth]{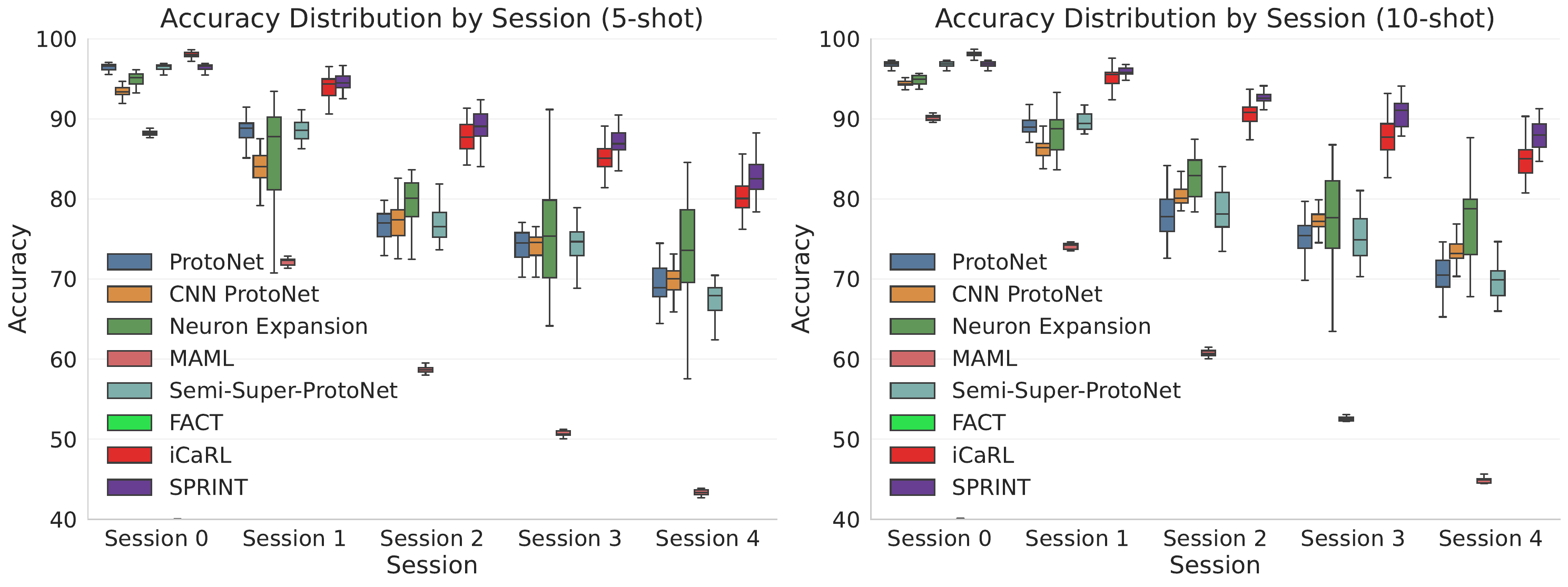}
    \caption{\textbf{Session-wise Performance Distribution and Stability.} Comparison of Top-1 accuracy distributions across 30 independent runs on the CICIDS2017 dataset for (left) 5-shot and (right) 10-shot settings. The boxplots visualize the variance and median performance at each incremental session. SPRINT (purple) demonstrates superior stability with minimal variance and negligible performance decay compared to baselines like iCaRL and MAML, which exhibit significant inter-run volatility and catastrophic forgetting.}
    \label{fig:session_boxplots}
\end{figure}

\begin{figure}[t]
    \centering
    \includegraphics[width=\textwidth]{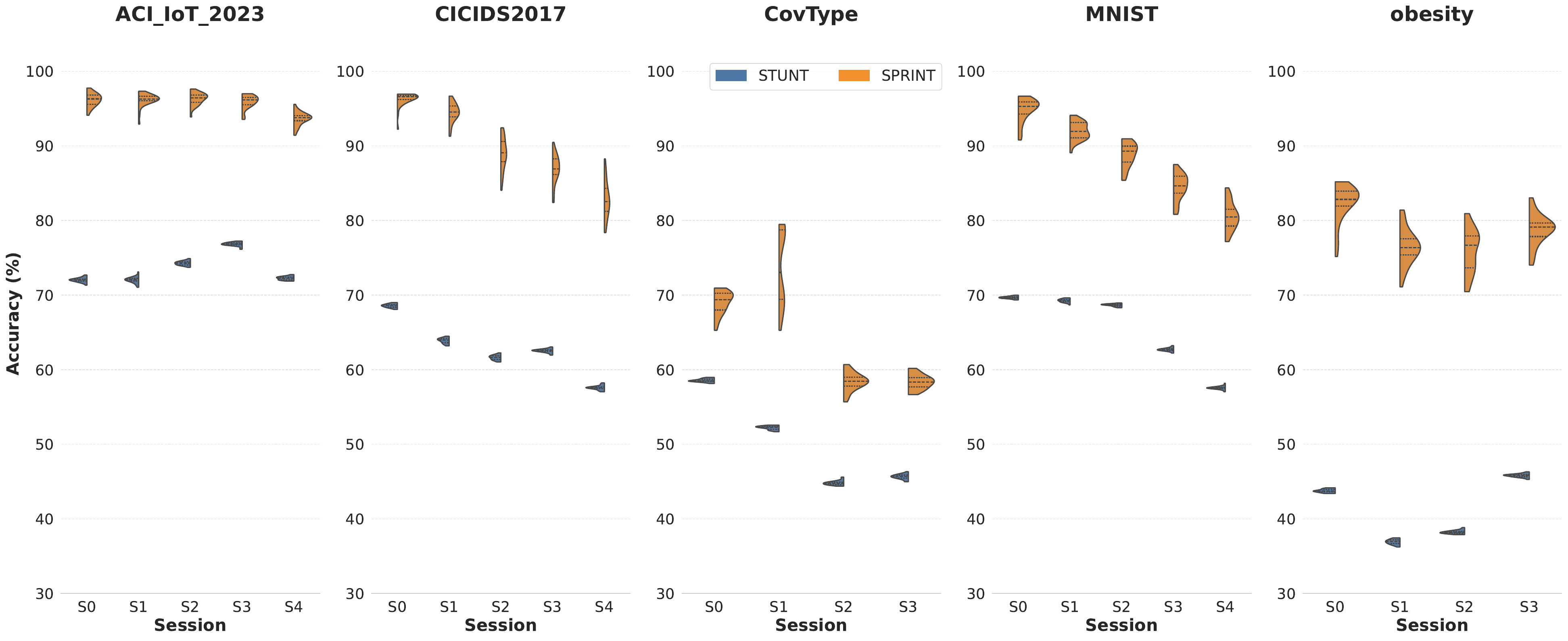}
    \caption{\textbf{Distribution of Incremental Performance (5-Shot).} Comparative violin plots of session-wise accuracy for STUNT \cite{nam2023stunt} and SPRINT across five datasets. The distributions reveal a distinct performance gap: SPRINT consistently operates in a high-accuracy regime (top quartiles), whereas STUNT remains stagnated at lower accuracy levels.}
    \label{fig:stunt_session}
\end{figure}

\textbf{Cross-Dataset Consistency.}
To verify generalization, we extended this analysis to all six benchmark datasets, specifically comparing SPRINT against the most competitive baseline, iCaRL. As summarized in Table \ref{tab:significance_summary}, SPRINT achieves statistically significant gains in 5 out of 6 datasets. The effect sizes range from \textbf{1.11} (CIC-IDS2017) to \textbf{6.96} (MNIST), indicating a robust improvement. The sole exception is CIC-IoT-2023, where SPRINT performs comparably to iCaRL (no significant gain detected), likely due to the specific feature distribution of that dataset favoring the replay mechanism of iCaRL. However, across the majority of diverse network traffic and tabular benchmarks, SPRINT consistently provides a statistically significant advantage.

\subsection{Detailed Stability and Variance Analysis}\label{app:stability_analysis} 

To evaluate the reliability of SPRINT against stochastic variations in episode generation and weight initialization, we analyze the performance distributions over 30 independent runs, as illustrated in Figure \ref{fig:session_boxplots}.

\textbf{Variance.} A critical requirement for deployment in security-critical domains is predictability. As shown in the boxplots, SPRINT exhibits markedly lower variance compared to competing methods. In the 5-shot setting, the Interquartile Range (IQR) for SPRINT remains tightly constrained even in the final session (Session 4), whereas baselines like iCaRL and CNN-ProtoNet show expanded whiskers and broader IQRs. This indicates that SPRINT's dual-loss optimization landscape is robust to random seed variations, consistently converging to high-quality solutions regardless of the specific support set sampled.

\textbf{Mitigation of Catastrophic Forgetting.} The session-wise trajectory reveals distinct failure modes in baseline architectures. Gradient-based meta-learners like MAML suffer from severe degradation, with median accuracy dropping precipitously after the first incremental session. Similarly, while FACT maintains reasonable stability initially, it fails to scale to later sessions. In contrast, SPRINT maintains a near-flat performance curve. The median accuracy in the 5-shot setting remains above \textbf{93\%} throughout all sessions, with the lower quartile of the final session still outperforming the upper quartile of the strongest baseline (iCaRL).

\textbf{Sensitivity to Support Set Size.} Comparing the 5-shot (left) and 10-shot (right) panels demonstrates SPRINT's data efficiency. While increasing the shot count to $K=10$ naturally tightens the distributions for all models, SPRINT's relative advantage persists. Even with abundant data, replay-based methods like iCaRL struggle to balance the plasticity-stability trade-off as effectively as SPRINT, which leverages high-confidence pseudo-labels to virtually expand the support set, effectively simulating a higher-shot regime even when data is scarce.

\subsection{Additional Baseline Comparisons: Self-Supervised Learning (STUNT)}
\label{app:stunt_results}

\begin{table}[h]
\centering
\caption{Comparison of self-supervised baseline STUNT versus SPRINT on 5-shot incremental settings. While STUNT exhibits low PD in some cases due to poor initial base learning, SPRINT consistently achieves superior final accuracy ($A_{last}$) while maintaining low performance degradation.}
\label{tab:stunt_vs_sprint}
\begin{tabular}{l|cc|cc}
\toprule
\multirow{2}{*}{\textbf{Dataset}} & \multicolumn{2}{c|}{\textbf{STUNT}} & \multicolumn{2}{c}{\textbf{SPRINT}} \\
 & \textbf{PD (\%)} $\downarrow$ & \textbf{Acc (\%)} $\uparrow$ & \textbf{PD (\%)} $\downarrow$ & \textbf{Acc (\%)} $\uparrow$ \\
\midrule
\textbf{ACI-IoT-2023} & -0.23 & 72.26 & {2.54} & {93.63} \\
\textbf{CIC-IoT-2023} & -3.42 & 54.95 & {-3.94} & 65.81 \\
\textbf{CICIDS2017} & 10.99 & 57.58 & {13.45} & {82.80} \\
\textbf{CovType} & 12.91 & 45.65 & 10.62 & {58.36} \\
\textbf{MNIST} & 12.11 & 57.56 & 5.18 & {84.85} \\
\midrule
\textbf{Average} & 6.47 & 57.60 & \textbf{5.24} & \textbf{77.37} \\
\bottomrule
\end{tabular}%

\end{table}

To ensure a comprehensive evaluation, we extended our study to include \textbf{STUNT} (Few-shot Tabular Learning with Self-generated Tasks) \cite{nam2023stunt}, a leading self-supervised method for tabular few-shot learning. The comparative results against SPRINT are detailed in Table \ref{tab:stunt_vs_sprint}.

\textbf{Analysis of the Stability-Plasticity Trade-off.} While STUNT achieves low, and occasionally \textbf{negative}, Performance Dropping (PD) rates (e.g., -3.42\% on CIC-IoT-2023), this metric presents a deceptive picture of stability. \textbf{Although negative PD typically implies backward transfer (performance gain), here it is an artifact of representation rigidity and a performance floor effect.} As evidenced by the absolute accuracy ($A_{last}$), STUNT significantly underperforms supervised and semi-supervised baselines. For instance, on CICIDS2017, STUNT achieves a final accuracy of only 57.58\% compared to SPRINT's 82.80\%.

This phenomenon indicates that STUNT suffers from a \textit{failure to adapt}: the model initializes with low discriminative power (preventing significant degradation) and remains trapped in that sub-optimal regime throughout incremental sessions. Unlike SPRINT, which demonstrates high plasticity (learning new classes effectively) while maintaining stability via dual-loss optimization, STUNT prioritizes representation rigidity at the expense of acquiring novel concepts.
Figure \ref{fig:stunt_session} corroborates this, showing that STUNT's accuracy distributions are consistently suppressed in lower quartiles compared to the high-confidence distributions of SPRINT. Consequently, we excluded these results from the main text to focus the discussion on baselines that achieve operationally viable accuracy.

%%%%%%%%%%%%%%%%%%%%%%%%%%%%%%%%%%%%%%%%%%%%%%%%%%%%%%%%%%%%%%%%%%%%%%%%%%%%%%%
%%%%%%%%%%%%%%%%%%%%%%%%%%%%%%%%%%%%%%%%%%%%%%%%%%%%%%%%%%%%%%%%%%%%%%%%%%%%%%%

\end{document}